\documentclass[sigconf]{acmart}
\usepackage{subfigure}
\usepackage{enumitem}
\usepackage{bm}
\usepackage{multirow}

\usepackage{lipsum}
\usepackage{amsthm}

\usepackage{algorithm}
\usepackage[noend]{algpseudocode}

\renewcommand\footnotetextcopyrightpermission[1]{}
\newtheoremstyle{mystyle}%
{\topsep}{\topsep}
{}{}
{\bfseries}{.}
{ }{\thmname{#1}\thmnumber{ #2} \thmnote{(#3)}}
\theoremstyle{mystyle}
\newtheorem{definition}{Definition}
\newtheorem{problem}{Problem}

\newcommand{\specialcell}[2][c]{%
	\begin{tabular}[#1]{@{}c@{}}#2\end{tabular}}

\AtBeginDocument{%
	\providecommand\BibTeX{{%
			\normalfont B\kern-0.5em{\scshape i\kern-0.25em b}\kern-0.8em\TeX}}}

\begin{document}

	\title{Knowledge-infused Contrastive Learning for Urban Imagery-based Socioeconomic Prediction}

	\author{Yu Liu}
	\affiliation{%
		\institution{Tsinghua University}
		\city{Beijing}
		\country{China}
	}
	\email{liuyu2419@126.com}

	\author{Xin Zhang}
	\affiliation{%
		\institution{Tsinghua University}
		\city{Beijing}
		\country{China}
	}
	\email{zhangxin4087@163.com}

	\author{Jingtao Ding}
	\authornote{Corresponding author.}
	\affiliation{%
		\institution{Tsinghua University}
		\city{Beijing}
		\country{China}
	}
	\email{dingjt15@tsinghua.org.cn}

	\author{Yanxin Xi}
	\affiliation{%
		\institution{University of Helsinki}
		\city{Finland}
		\country{China}
	}
	\email{yanxin.xi@helsinki.fi}
	
	\author{Yong Li}
	\affiliation{
		\institution{Tsinghua University}
		\city{Beijing}
		\country{China}
	}
	\email{liyong07@tsinghua.edu.cn}
	
	\begin{abstract}
		Monitoring sustainable development goals requires accurate and timely socioeconomic statistics, while ubiquitous and frequently-updated urban imagery in web like satellite/street view images has emerged as an important source for socioeconomic prediction. Especially, recent studies turn to self-supervised contrastive learning with manually designed similarity metrics for urban imagery representation learning and further socioeconomic prediction, which however suffers from effectiveness and robustness issues. To address such issues, in this paper, we propose a \underline{Know}ledge-infused \underline{C}ontrastive \underline{L}earning (KnowCL) model for urban imagery-based socioeconomic prediction. Specifically, we firstly introduce knowledge graph (KG) to effectively model the urban knowledge in spatiality, mobility, etc., and then build neural network based encoders to learn representations of an urban image in associated semantic and visual spaces, respectively. Finally, we design a cross-modality based contrastive learning framework with a novel image-KG contrastive loss, which maximizes the mutual information between semantic and visual representations for knowledge infusion. Extensive experiments of applying the learnt visual representations for socioeconomic prediction on three datasets demonstrate the superior performance of KnowCL with over 30\% improvements on $R^2$ compared with baselines. Especially, our proposed KnowCL model can apply to both satellite and street imagery with both effectiveness and transferability achieved, which provides insights into urban imagery-based socioeconomic prediction.
	\end{abstract}


	\maketitle
	\section{Introduction}\label{sec:intro}
	Driven by the rapid urbanization, more than half of the world population-4.4 billion inhabitants-live in cities and contribute over 80\% of global GDP today \citep{urban_development}, which makes cities an increasingly important role in achieving United Nations Sustainable Development Goals (SDGs) on economy, education, environment, health, etc. \citep{un2018world,un2022sdg}. Especially, socioeconomic indicators like population, educational background and household income are good proxies for SDG monitoring \cite{gebru2017using}. The traditional door-to-door surveys for such statistics however are costly, labor-intensive, time-consuming and further affected by recent COVID-19 pandemic \citep{un2022sdg}. In contrast, the inclusive, ubiquitous and frequently-updated web applications paves the way for high-quality, economical and timely SDG monitoring. Recently, researchers predict socioeconomic indicators with the enormous amount of web data especially the urban imagery \cite{burke2021using,li2022predicting,wang2018urban}, i.e., the satellite imagery and the street view imagery provided in web map services like Google Map and web platforms like Instagram. 
	
	\begin{figure}[htbp]
		\centering
		\includegraphics[width=.9\linewidth]{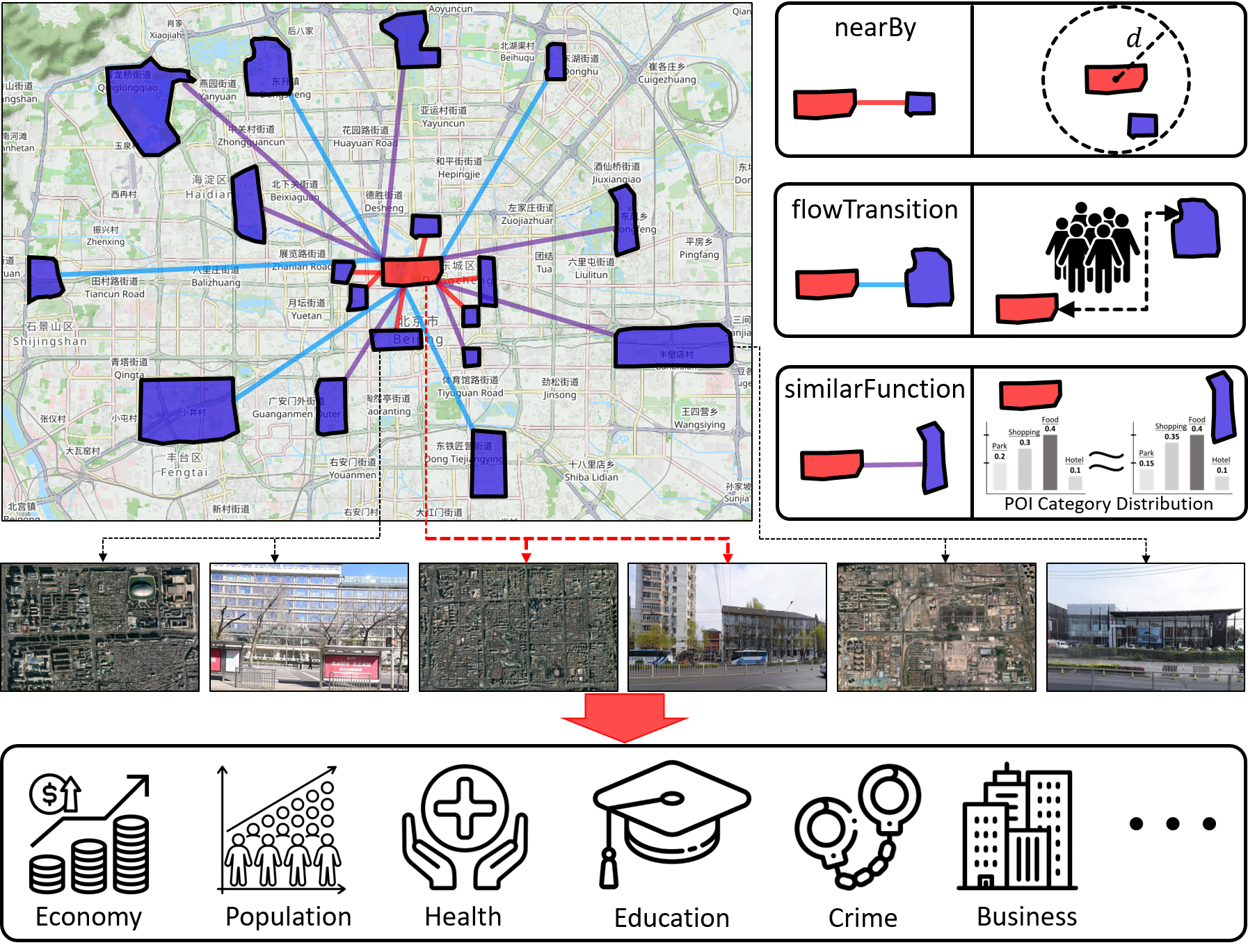}
		\vspace{-10px}
		\caption{Illustration of infusing various types of knowledge for urban imagery-based socioeconomic prediction, e.g., ``nearBy'', ``flowTransition'' and ``similarFunction'' relational links describe urban knowledge in spatiality, mobility and function (POI category distribution), respectively.}
		\label{fig:idea}
		\vspace{-10px}
	\end{figure}

	Built upon the great success of deep learning in computer vision \citep{deng2009imagenet,krizhevsky2017imagenet}, most studies adopt convolutional neural networks (CNNs) to learn visual representations of urban imagery for socioeconomic prediction. Specifically, earlier studies follow the task-specific supervised learning for visual representations with neighborhood demographics \citep{gebru2017using,abitbol2020interpretable} and country poverty \citep{jean2016combining,yeh2020using} as supervision signals, which require massive labeled data for training and suffer from generalization issues \citep{rolf2021generalizable}. To overcome such issues, recent studies turn to self-supervised learning with contrastive objectives, a.k.a, contrastive learning \citep{liu2021self}, and learn a single representation vector for one image to generalize across diverse prediction tasks \citep{wang2022self,ahn2020teaching}. Based on manually designed similarity metrics, these studies learn visual representations of urban imagery by maximizing the agreement between similar images in latent space under an image view-based contrastive learning framework \citep{zhang2020contrastive}. For example, a commonly used metric on spatiality knowledge is from Tobler's First Law of Geography \citep{miller2004tobler} that spatially near images should have similar semantics and thus closer representations \citep{jean2019tile2vec,wang2020urban2vec,kang2020deep,bjorck2021accelerating}. Moreover, a recent study \citep{xi2022beyond} applies the typical contrastive learning framework, SimCLR \citep{chen2020simple}, for satellite imagery-based socioeconomic prediction, assuming that images with similar point of interest (POI) features should have closer representations \footnote{Here POI features correspond to POI category distributions of regions identified in urban imagery.}. Owing to the task-agnostic representation learning from unlabeled data, contrastive learning becomes a promising avenue for urban imagery-based socioeconomic prediction.

	Despite this, existing contrastive learning based methods heavily rely on manually designed similarity metrics for urban imagery representation learning, which only capture one or two types of semantic knowledge in urban environment and thus affect the practical performance in socioeconomic prediction. According to recent works of leveraging multi-source urban data for socioeconomic prediction \citep{zhang2021multi,wu2022multi}, there are various types of semantic knowledge available for urban imagery-based socioeconomic prediction, e.g., the spatiality knowledge of spatial neighborhood, the mobility knowledge of significant flow transitions and the function knowledge of similar POI category distributions, as shown in Figure~\ref{fig:idea}. Thus, how to infuse comprehensive knowledge into contrastive learning for urban imagery-based socioeconomic prediction becomes an important research problem, which however is challenging in:
	\begin{itemize}[leftmargin=10px]
		\item \textbf{Effective structure for knowledge identification.} Unlike the well-known domain knowledge like Tobler's First Law of Geography, other types of aforementioned knowledge lack explicit domain definition. Moreover, further knowledge infusion requires an effective structure to store and represent the knowledge, increasing the difficulty of knowledge identification for urban imagery-based socioeconomic prediction. 
		\item \textbf{Contrastive learning for knowledge infusion.} Existing studies adopt the image view-based contrastive learning framework where the similarity metric is manually designed with a single type of knowledge. Therefore, they fail to infuse various types of knowledge for urban imagery-based socioeconomic prediction.
	\end{itemize}
	
	To address such challenges, in this paper, we propose a \underline{Know}ledge-infused \underline{C}ontrastive \underline{L}earning model for urban imagery-based socioeconomic prediction, termed as KnowCL. Firstly, motivated by the recent success of the structured knowledge graph (KG) for urban knowledge modeling \citep{zhuang2017understanding,liu2022developing,liu2021urban,wang2021spatio,wang2020incremental}, we introduce urban knowledge graph (UrbanKG) to identify comprehensive knowledge in multi-source urban data. In the UrbanKG, entity nodes characterize urban elements like regions, POIs and business centers, while relation edges describe semantic connections between them, i.e., the knowledge in spatiality, mobility and function. Moreover, we present cross-modality based contrastive learning for knowledge infusion by exploiting the naturally associated pairing of urban imagery and regions in UrbanKG\footnote{The knowledge graph is identified as another kind of modality data versus urban imagery data in visual modality.}. To be specific, for an urban image, KnowCL develops a visual encoder to extract its visual representation, and a semantic encoder to extract KG embedding of its associated region entity in UrbanKG \citep{wang2017knowledge}, which are further optimized for maximum agreement with contrastive loss on image-KG pairs. The learnt visual representations of urban imagery are further fed into traditional regression models for diverse socioeconomic prediction tasks. Therefore, KnowCL firstly leverages UrbanKG for knowledge identification, then represents comprehensive knowledge with KG embedding, and combines with cross-modality based contrastive learning to achieve knowledge infusion for urban imagery-based socioeconomic prediction. The main contributions of this paper are summarized as follows:
	\begin{itemize}[leftmargin=10px]
		\item To the best of our knowledge, we are the first to investigate KG for urban imagery-based socioeconomic prediction, which provides an effective structure to comprehensively identify the semantic knowledge in spatiality, mobility, function, etc.
		\item We propose a cross-modality based contrastive learning framework, which infuses semantic knowledge into visual representations of urban imagery via the novel contrastive objective between the image and KG modalities. The proposed framework might shed light on urban imagery representation learning.
		\item We conduct extensive experiments on three cities of Beijing, Shanghai and New York across six socioeconomic indicators. The results on both satellite and street view imagery demonstrate that our proposed framework achieves significant performance improvement compared with state-of-the-art models. Further ablation studies and analysis confirm the effectiveness and transferability of knowledge infusion for urban imagery-based socioeconomic prediction. 
	\end{itemize}

	\section{Related Work}\label{sec:related_work}
	As described before, the urban imagery-based socioeconomic prediction studies focus on urban imagery representation learning, which extracts visual representations for downstream socioeconomic prediction tasks. Based on whether the urban imagery representation learning process needs supervision signals from downstream tasks, related studies can be classified into supervised learning, unsupervised learning and self-supervised learning\footnote{The self-supervised learning is separated from the unsupervised one, which emphasizes using supervision signals generated from data itself.}.
	
	\textbf{Supervised Urban Imagery Representation Learning for Socioeconomic Prediction.} We first discuss about the input source of satellite imagery. With the CNN model pre-trained on ImageNet \citep{deng2009imagenet} and light intensity as supervision signal, both Jean \emph{et al.} \citep{jean2016combining} and Yeh \emph{et al.} \citep{yeh2020using} extract satellite imagery representations for assets prediction in Africa. Similar frameworks are proposed in \citep{he2018perceiving, park2022learning} for economic indicator prediction. Han \emph{et al.} \citep{han2020lightweight} train a teacher-student network with limited labels to predict demographics like household and income. As for street imagery case, Gerbu \emph{et al.} \citep{gebru2017using} train a CNN model to identify the types and number of cars in street view imagery, which are further used to estimate socioeconomic indicators like race and education. Lee \emph{et al.} \citep{lee2021predicting} leverage semantic segmentation and graph convolution network (GCN) to predict livelihood indicators of wealth index and BMI. Moreover, Law \emph{et al.} \citep{law2019take} extract features from both satellite and street view imagery to estimate the house prices. However, above studies learn urban imagery representations supervised by a specific downstream task, i.e., the learnt representations cannot generalize to various socioeconomic prediction tasks. 	 
	
	\textbf{Unsupervised Urban Imagery Representation Learning for Socioeconomic Prediction.} Han \emph{et al.} \citep{han2020learning} adopt clustering algorithm and partial order graph to distinguish economic development of satellite imagery, which are further used to train a scoring model for urbanization prediction. Suel \emph{et al.} \citep{suel2019measuring} apply the pre-trained CNN model to extract street view imagery representations for inequality measurement in urban environment. Besides, He \emph{et al.} \citep{he2018perceiving} extract traditional image features like histogram of oriented gradients from both satellite and street view imagery to predict commercial activeness. However, such unsupervised learning methods only capture shallow features of urban imagery, which lead to inferior performance.
		
	\textbf{Self-supervised Urban Imagery Representation Learning for Socioeconomic Prediction.} Motivated by the milestones of self-supervised learning achieved in computer vision \citep{wang2022self,chen2020simple,jing2020self}, researchers also leverage self-supervised learning especially contrastive learning for urban imagery-based socioeconomic prediction, and focus on similarity metric design to distill expressive representations of urban imagery. Especially, most studies follow the Tobler's First Law of Geography \citep{miller2004tobler} that ``everything is related to everything else, but near things are more related than distant things'', and design corresponding similarity metrics or loss forms. For example, Jean \emph{et al.} \citep{jean2019tile2vec} employs the triplet loss to minimize the distance between representations of spatially near satellite images but maximize the distance between those of spatially distant pairs, while Wang \emph{et al.} \citep{wang2020urban2vec} employs the similar loss for street view imagery case. Moreover, recent studies \citep{kang2020deep,bjorck2021accelerating,xi2022beyond} mainly adopt the InfoNCE loss \citep{oord2018representation} with SimCLR framework \citep{chen2020simple} to encode such spatiality knowledge into visual representations. Xi \emph{et al.} \citep{xi2022beyond} further incorporate a POI-based similarity metric such that images corresponding to similar POI category distributions should have closer visual representations. Furthermore, Li \emph{et al.} \citep{li2022predicting} consider the spatiality based similarity metrics for both satellite imagery and street imagery. According to the discussion above, most existing studies adopt the image view-based contrastive learning framework, where a pair of images are compared to capture spatiality knowledge, failing to infuse various types of knowledge together. In comparison, our proposed KnowCL model captures comprehensive knowledge via UrbanKG and achieves effective knowledge infusion with cross-modality based contrastive learning framework.

	\section{Preliminaries \& Problem Statement} \label{sec:pre}	
	As stated in \emph{The sustainable Development Goals Report} \cite{un2022sdg}, the SDGs determine the survival of humanity, facing the current confluence of crises. Especially, various socioeconomic indicators are characterized for SDG monitoring \citep{corsi2012demographic}:	 
	\begin{definition}[Socioeconomic Indicator]
		Socioeconomic indicators measure the status of the region/nation on the socioeconomic scale, determined by a combination of social and economic factors such as population, amount and kind of education, household consumption, crime rate, etc. 
	\end{definition}
	
	Moreover, the urban region becomes an important subject for socioeconomic indicator investigation, which are defined as:
	\begin{definition}[Urban Region]
		A city can be partitioned into a set of urban regions $\mathcal{A}$, following certain partition criteria like road network division and administrative division \citep{wang2020urban2vec,han2020lightweight}.
	\end{definition}
	
	The urban imagery includes the satellite imagery and the street view imagery, which are visual appearances of the city from overhead view and ground-level view, respectively \citep{wang2018urban,li2022predicting}. Figure~\ref{fig:idea} provides some examples of the urban imagery. Specifically, the satellite images are collected by satellites, which capture the structure of regions in the city. The street view images are taken by automobiles or citizens along the street, which capture the internal environment of regions in the city. Moreover, an urban region usually associates with one satellite image but multiple street view images taken at different locations therein, which are defined as:
	\begin{definition}[Urban Imagery]
		Given a city, the urban imagery set is denoted by  $\mathcal{I}=\mathcal{I}^{\text{SI}}/\mathcal{I}^{\text{SV}}$ with the satellite imagery set $\mathcal{I}^{\text{SI}}$ and the street view imagery set $\mathcal{I}^{\text{SV}}$. For $\forall a\in\mathcal{A}$, its associated satellite image is denoted by $I^{\text{SI}}_{a}\in \mathcal{I}^{\text{SI}}$, and its associated $n$ street view images are denoted by $I^{\text{SV}}_{a}=\{I^{\text{SV}}_{a,1},\!\cdots\!,I^{\text{SV}}_{a,n}\}$ with $I^{\text{SV}}_{a,1},\!\cdots\!,I^{\text{SV}}_{a,n} \in \mathcal{I}^{\text{SV}}$.
	\end{definition}
	
	The UrbanKG generalizes the commonly used KG concept \citep{wang2017knowledge,hogan2021knowledge} to urban domain for urban knowledge modeling \citep{zhuang2017understanding,liu2022developing,liu2021urban,wang2021spatio}, which is defined as:
	\begin{definition}[Urban Knowledge Graph] 
		An UrbanKG is defined as a multi-relational graph $\mathcal{G}=(\mathcal{E}, \mathcal{R}, \mathcal{F})$, where $\mathcal{E}$, $\mathcal{R}$ and $\mathcal{F}$ are the sets of entities, relations and facts, respectively, with $\mathcal{F}=\{(e_h,r,e_t)| e_h,e_t\in\mathcal{E}, r\in\mathcal{R}\}$ hold. Especially, the entity set $\mathcal{E}$ includes urban elements like regions, POIs, business centers and categories, while the relation set $\mathcal{R}$ describes their semantic connections on spatiality, mobility, function and business. The set of region entities in $\mathcal{G}$ corresponds to above defined region set $\mathcal{A}$. 
	\end{definition}
	 Details of the UrbanKG will be presented in detail in Section~\ref{sec:urbankg}. To capture the semantic knowledge for downstream applications, recent studies learn to embed entities and relations of a KG into low-dimensional vector space, a.k.a., KG embeddings \citep{wang2017knowledge,hogan2021knowledge,vashishth2019composition}.
	 
	 Based on the preliminaries above, we then formally define the urban imagery-based socioeconomic prediction problem as follows.
	
	\begin{problem}[Urban Imagery-based Socioeconomic Prediction]
	Given the region set $\mathcal{A}$ with its associated urban imagery set $\mathcal{I}$, for $\forall a \in \mathcal{A}$, the main goal is to learn the visual representation $\bm{I}_a$ and well estimate the socioeconomic indicator $y_a$. The ground truth values of $y_a$ is assumed to be unknown in urban imagery representation learning. 
	\end{problem}

	\section{Methodology} \label{sec:model}
	\subsection{Framework Overview}
\begin{figure}[htbp]
		\vspace{-5px}
		\centering
		\includegraphics[width=.9\linewidth]{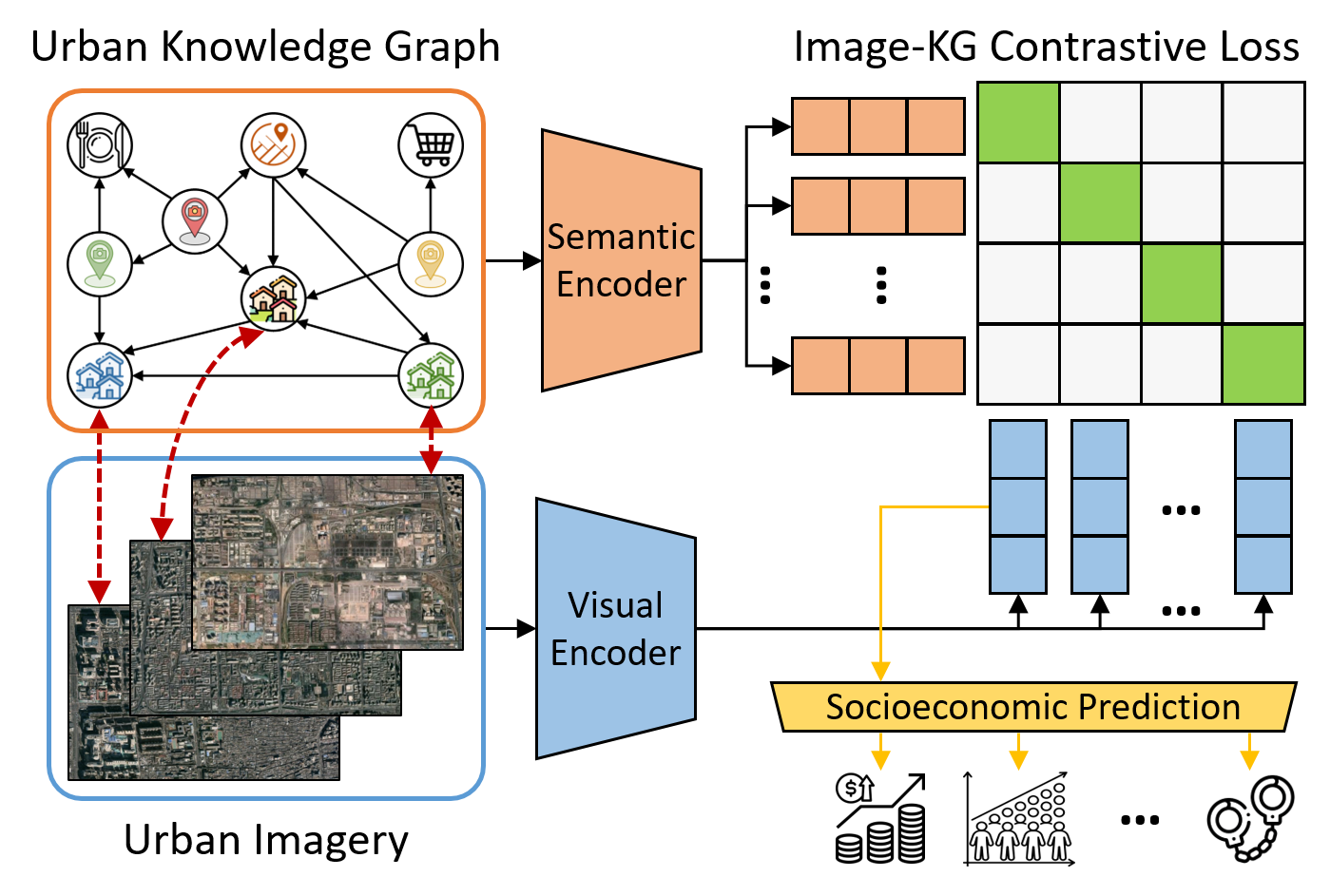}
		\vspace{-10px}
		\caption{The main framework of KnowCL model, where the urban imagery input can be either satellite imagery or street view imagery. The projector heads after encoders are omitted for simplicity in the illustration.}
		\label{fig:framework}
		\vspace{-5px}
	\end{figure}

	Figure~\ref{fig:framework} presents the main framework of our proposed KnowCL model for urban imagery-based socioeconomic prediction problem. Since we consider the socioeconomic indicators on region level, we solve the challenges of knowledge identification and knowledge infusion with focus on regions in the city. 
		
	To identify the comprehensive knowledge in urban environment, we firstly introduce the recently proposed UrbanKG structure to store and represent urban knowledge related with regions, which is then fed into a GCN-based semantic encoder to learn KG embeddings for region entities therein. As for satellite/street view images associated with regions, a CNN-based visual encoder is adopted for visual representations. Furthermore, we propose cross-modality based contrastive learning framework to achieve knowledge infusion. Especially, the designed image-KG contrastive loss encourages one region's KG embedding and its associated urban imagery representation to exhibit high mutual information, through which the semantic knowledge preserved in KG embedding is successfully infused into urban imagery representation. Finally, the knowledge-infused urban imagery representations are leveraged for diverse socioeconomic prediction tasks.
	
	\subsection{Urban Knowledge Identification}
	As defined in Section~\ref{sec:pre}, we introduce UrbanKG to identify the urban knowledge for socioeconomic prediction. Specifically, the entities in UrbanKG include regions partitioned by road network, business centers of commercial and consumption activities, POIs of infrastructures like restaurants, markets and schools, as well as categories of POI attributes, e.g., food, shopping, education, etc. Thus, the entity types in UrbanKG are Region, Business Center (BC), POI and Category. 
	
	\begin{table}[htbp]
		\vspace{-5px}
		\caption{The captured knowledge and corresponding relational structures in UrbanKG.}\label{tab:relations}
		\vspace{-10px}
 \def\arraystretch{0.95}%
		\begin{tabular}{c|c|c|c}
			\toprule
\textbf{Knowledge} &	\textbf{Relation} &  \textbf{Head Entity} & \textbf{Tail Entity}  \\
			\midrule
\multirow{3}{*}{Spatiality} &	\emph{borderBy} & Region & Region  \\
		&	\emph{nearBy} &  Region &  Region    \\
		&	\emph{locateAt}  & POI & Region \\
		\midrule
\multirow{1}{*}{Mobility}	& \emph{flowTransition}  & Region & Region \\
	\midrule
\multirow{3}{*}{Function} &	\emph{similarFunction} & Region & Region \\
		&	\emph{coCheckin}  & POI & POI    \\
		&	\emph{cateOf}  & POI & Category \\
		\midrule
\multirow{3}{*}{Business} & \emph{provideService}  & BC & Region \\
		&	\emph{belongTo} & POI & BC   \\
	&	\emph{competitive}  & POI & POI \\
			\bottomrule
		\end{tabular}
	\end{table}

	Moreover, we model the comprehensive knowledge in multi-source urban data as semantic relations, which are summarized in Table~\ref{tab:relations}. Various types of knowledge are represented in triple form with relation, head entity and tail entity. 
	\begin{itemize}[leftmargin=10px]
		\item \textbf{Spatiality.} We determine \emph{borderBy} and \emph{nearBy} relational links by spatial distance between regions, and use \emph{locateAt} to identify POIs' spatially located regions. 
		\item \textbf{Mobility.} we aggregate individual mobility trajectories to induce the significant flow transition between regions, which are linked by \emph{flowTransition}.
		\item \textbf{Function.} We consider widely used features in urban computing tasks \citep{zheng2014urban} as function knowledge. For example, \emph{coCheckin} connects highly correlated POIs in terms of concurrence in check-in data, which implies two POIs are consecutively visited by several people, e.g., a cinema and a neighboring restaurant \citep{liu2021knowledge}. Since the region function is usually featured by POI category distribution therein \citep{xi2022beyond}, we connect regions with similar POI category distribution via \emph{similarFunction}. Besides, \emph{cateOf} describes the category attribute of POIs.
		\item \textbf{Business.} We connect regions with their neighboring business centers via \emph{provideService}, to capture the economic status of regions. Similarly, POIs are connected with neighboring business centers via \emph{belongTo}. To further identify the competitiveness between POIs, neighboring POIs with the same category are connected via \emph{competitive} \citep{li2020competitive}.    
	\end{itemize}  
	Besides, reverse edges are added to model inverse relations, e.g., (POI, \emph{locateAt}, Region) and (Region, \emph{$\sim$locateAt}, POI). Following the structure above, we construct the UrbanKG with urban knowledge identified for socioeconomic prediction. The construction details can be referred to Section~\ref{sec:add_data}.
	
	\subsection{Encoder Design} \label{sec:urbankg}
	\subsubsection{Semantic Encoder Design}
	The semantic encoder aims to learn region embeddings with urban knowledge represented. To fully exploit both semantic information of various relations and structural information of graph topology in UrbanKG, we adopt the GCN-based encoder for region embeddings \citep{vashishth2019composition,liu2021knowledge}. 
	
	Given the UrbanKG $\mathcal{G}=(\mathcal{E}, \mathcal{R}, \mathcal{F})$, for $\forall v\in\mathcal{E}, r\in\mathcal{R}$, their $d$-dimensional embeddings after $l$ layer are denoted by $\bm{e}^{l+1}_v$ and $\bm{r}^{l+1}$, respectively. The neighborhood of $v$ is denoted by  $\mathcal{N}_v=\{(u,r)\vert(u,r,v)\in\mathcal{F}\}$, and $\bm{e}^{l+1}_v\in\mathbb{R}^d$ can be calculated as:
	\setlength{\belowdisplayskip}{0pt} \setlength{\belowdisplayshortskip}{0pt}
	\setlength{\abovedisplayskip}{0pt} \setlength{\abovedisplayshortskip}{0pt}
	\begin{align}
		\bm{e}^{l+1}_v=\sigma\left( \sum_{(u,r)\in\mathcal{N}_v}  \bm{W}^l_{\text{dir}(r)} \phi(\bm{e}^l_u, \bm{r}^l) + \bm{W}^l_{\text{self}}\bm{e}^l_v\right),
	\end{align}

	where $\bm{W}^l_{\text{dir(r)}}$ and $\bm{W}^l_{\text{self}}$ are direction-specific projection matrices for incoming/outgoing relations and self loop relation, respectively. $\phi:\mathbb{R}^d\times\mathbb{R}^d\rightarrow\mathbb{R}^d$ is the composition function for message calculation in GCN, e.g., element-wise summation and element-wise product \citep{vashishth2019composition}. $\sigma(\cdot)$ is an activation function. Besides, we use pre-trained embeddings from TuckER \citep{balazevic2019tucker} for initialized embeddings. 
	
	Let $f^{\text{KG}}(\cdot)$ denote the semantic encoder with $L$ layers of GCN following above design, and the region embedding for $a\in\mathcal{A}$ can be calculated as  $\bm{e}_a=f^{\text{KG}}(\mathcal{G}, a)$, i.e., $\bm{e}_a=\bm{e}^L_{a}$.

	\subsubsection{Visual Encoder Design}
	Our proposed KnowCL model allows various choices of network architectures for visual encoder design. For simplicity, we adopt the commonly used ResNet \citep{he2016deep} to obtain visual representations of urban imagery. 
	
	For $a\in\mathcal{A}$ with satellite imagery $I^{\text{SI}}_a$, the visual representation can be calculated as $\bm{I}^{\text{SI}}_a=\text{ResNet}(I^{\text{SI}}_a)$. As for  street view imagery $I^{\text{SV}}_{a}=\{I^{\text{SV}}_{a,1},\!\cdots\!,I^{\text{SV}}_{a,n}\}$, the visual representation is calculated by average pooling on street view images therein: 
	\begin{align}
	\bm{I}^{\text{SV}}_a=\frac{1}{n} \sum_{i=1}^n \text{ResNet}(I^{\text{SV}}_{a,i}).
	\end{align}
	
	Thus, let $f^{\text{Image}}(\cdot)$ denote the visual encoder designed above, and the urban imagery representation can be obtained by $\bm{I}_a=f^{\text{Image}}(I_a)$ with $I_a=I^{\text{SI}}_a/I^{\text{SV}}_a$ and $\bm{I}_a=\bm{I}^{\text{SI}}_a/\bm{I}^{\text{SV}}_a$.

	\subsection{Contrastive Loss Design \& Optimization}
	Motivated by cross-modality based contrastive learning between image and text modalities \citep{zhang2020contrastive,radford2021learning}, we design a novel image-KG contrastive loss for knowledge infusion. The core insight here is that both semantic representation (KG embedding) and visual representation (urban imagery representation) of a region should be close to each other.  
	
	First, for better representation quality, we introduce two independent projection heads $g^{\text{KG}}(\cdot)$ and $g^{\text{Image}}(\cdot)$ after semantic encoder and visual encoder, respectively, as validated in empirical studies \citep{chen2020simple,bjorck2021accelerating}. The corresponding outputs of $\tilde{\bm{e}}_a$ and $\tilde{\bm{I}}_a$ for $a\in\mathcal{A}$ can be expressed as:
	\begin{align}
		& \tilde{\bm{e}}_a = g^{\text{KG}}(\bm{e}_a) = \bm{W}^{\text{KG}}_2\text{ReLU}(\bm{W}^{\text{KG}}_1\bm{e}_a) \\
		& \tilde{\bm{I}}_a = g^{\text{Image}}(\bm{I}_a) = \bm{W}^{\text{Image}}_2\text{ReLU}(\bm{W}^{\text{Image}}_1\bm{I}_a),
	\end{align}
	where four projection matrices are used to project representations for both modalities from their encoder space to the same space for contrastive learning. 
	
	Moreover, following the core insight above, we extend the traditional InfoNCE loss \citep{oord2018representation} to image-KG contrastive loss, and the loss function for $a\in\mathcal{A}$ is expressed as:
	\begin{align}
	\mathcal{L}_a & = \mathcal{L}_a^{\text{Image}\rightarrow\text{KG}} + \mathcal{L}_a^{\text{KG}\rightarrow\text{Image}} \notag\\
		& = -\log\!\frac{\exp(\text{sim}(\tilde{\bm{I}}_a, \tilde{\bm{e}}_a))}{\sum_{i=1}^m \exp(\text{sim}(\tilde{\bm{I}}_a, \tilde{\bm{e}}_i))}\!-\! \log\!\frac{\exp(\text{sim}(\tilde{\bm{e}}_a,\tilde{\bm{I}}_a))}{\sum_{i=1}^m \exp(\text{sim}(\tilde{\bm{e}}_a, \tilde{\bm{I}}_i))}, \label{eq:loss}
	\end{align}
	where the loss is computed in a minibatch of $m$ samples, and $\text{sim}(\cdot)$ represents the inner product. $\mathcal{L}_a^{\text{Image}\rightarrow\text{KG}}$ and $\mathcal{L}_a^{\text{KG}\rightarrow\text{Image}}$ are image-to-KG and KG-to-image contrastive losses, respectively, which maximally preserve the mutual information between image-KG pairs. Unlike existing urban imagery-based socioeconomic prediction studies using image view-based contrastive loss in the same modality \citep{xi2022beyond,li2022predicting}, our proposed image-KG contrastive loss is based on cross modalities of inputs, which successfully infuses the comprehensive knowledge captured in region embeddings into urban imagery representations.
	
	By optimizing the image-KG contrastive loss on the whole data, we obtain knowledge-infused urban imagery representations $\bm{I}_{a\in\mathcal{A}}$ from the visual encoder, which are further fed into the regression module of multi-layer perceptron (MLP) for socioeconomic indicator training and prediction.

	\section{Experiments and Results}
	\subsection{Experimental Setup}
	\subsubsection{Datasets}
	We collect three datasets with urban imagery and socioeconomic indicator data for evaluation: Beijing (BJ), Shanghai (SH) and New York (NY). Regions in Beijing and Shanghai are partitioned by road network, while regions in New York are Census Block Groups (CBGs) used by US Census Bureau. The 256$\times$256-pixel satellite images with about 4.7 m-resolution are obtained from ArcGIS, which are further merged along irregular region boundaries for input satellite images. The 1024$\times$512-pixel street view images in Beijing and Shanghai as well as the 512$\times$512-pixel street view images in New York are obtained from Baidu Map API and Google Street API, respectively.
	
	As for socioeconomic indicator data, Beijing dataset includes \textbf{(1) Pop.}: population data from WorldPop, \textbf{(2) Econ.}: economic activity data from \citep{dong2021gridded}, \textbf{(3) Rest.}: restaurant business (takeaway order data) and \textbf{(4) Consp.}: consumption data from a life service platform, while Shanghai dataset includes \textbf{(1) Pop.}: population and \textbf{(2) Econ.}: economic activity data from same sources. New York dataset includes \textbf{(1) Pop.}: population and \textbf{(2) Edu.}: education data from SafeGraph, and \textbf{(3) Crime}: crime data from NYC Open Data. The UrbanKG data for three datasets are from \citep{liu2021knowledge,wang2020incremental}, and business knowledge related relations are omitted in New York dataset due to a lack of source data. All socioeconomic indicators are converted into logarithmic scale, i.e. $y=\ln(y_{\text{raw}}+1)$. Besides, regions in datasets with over 40 street view images are selected and randomly split into train/valid/test sets by a proportion of 6:2:2 in the socioeconomic prediction step. Table~\ref{tab:data} summarizes dataset statistics and details are provided in Section~\ref{sec:add_data}.

	\begin{table}[htbp]
		\vspace{-5px}
		\caption{Dataset Statistics.}\label{tab:data}
		\vspace{-10px}
		\def\arraystretch{0.95}
		\begin{tabular}{cccccccc}
			\toprule
			Dataset  & \#Region & \#SV & \#SI &  $|\mathcal{E}|$  & $|\mathcal{R}|$ & $|\mathcal{F}|$  \\
			\midrule
			Beijing  & 789 & 31,560 & 18,289 & 36,752 & 10 & 188,985   \\
			Shanghai  & 1,553 & 62,120 & 5,904 & 58,145  & 10 & 363,159 \\
			New York & 1,142 &  45,680 & 1,560 & 87,020 & 6 & 357,464  \\
			\bottomrule     
		\end{tabular}
		\vspace{-10px}
	\end{table}

	\begin{table*}[htbp]
		\caption{Satellite imagery-based socioeconomic prediction results on three datasets. Best results are in bold and the best results are underlined. The last row shows relative improvement in percentage.}
		\vspace{-10px}
		\setlength\tabcolsep{2.2pt}
		\label{tab:si_res}
		 \def\arraystretch{0.95}%
		\begin{tabular}{c|cccccccc|cccc|cccccc}
			\toprule
			\textbf{Dataset} & \multicolumn{8}{c|}{\textbf{Beijing}} & \multicolumn{4}{c|}{\textbf{Shanghai}} & \multicolumn{6}{c}{\textbf{New York}} \\
			\midrule
			\multirow{2}{*}{\textbf{Model}} & \multicolumn{2}{c}{\textbf{Pop.}} & \multicolumn{2}{c}{\specialcell{\textbf{Econ.}}} & \multicolumn{2}{c}{\specialcell{\textbf{Rest.}}} & \multicolumn{2}{c|}{{\textbf{Consp.}}} & \multicolumn{2}{c}{\textbf{Pop.}} & \multicolumn{2}{c|}{{\specialcell{\textbf{Econ.}}}} & \multicolumn{2}{c}{\textbf{Pop.}} & \multicolumn{2}{c}{\textbf{Edu.}} & \multicolumn{2}{c}{\textbf{Crime}} \\ 
			&  $R^2$ & RMSE & $R^2$ & RMSE & $R^2$ & RMSE & $R^2$ & RMSE & $R^2$ & RMSE & $R^2$ & RMSE & $R^2$ & RMSE & $R^2$ & RMSE & $R^2$ & RMSE \\
			\midrule
			\multicolumn{1}{c|}{ResNet-18} & 0.277 & 0.887  & 0.168  & 1.465 &  0.146 & 2.569 &  {0.125} & \multicolumn{1}{c|}{3.435} & 0.007 & 1.027 & {0.082} & \multicolumn{1}{c|}{1.674} & -0.404 & 0.788 & 0.518  & 0.085 &  0.324 & 0.751 \\
			
			\multicolumn{1}{c|}{Tile2Vec} & 0.274 & 0.888 & 0.092 & 1.531 & 0.108 & 2.626 & {0.074} & \multicolumn{1}{c|}{3.534} & 0.125 & 1.041 & {0.065} & 1.689 & \underline{0.143} & \underline{0.615} & 0.533 &  0.084 & 0.382 & 0.718 \\
			
			\multicolumn{1}{c|}{READ} & 0.300 & 0.872 & 0.173 & 1.461 & 0.222 & 2.451 & {0.213}  &  \multicolumn{1}{c|}{3.258} & 0.154 & 0.949 & {0.097} & 1.660 & -0.034 & 0.676 & 0.534 & 0.084 & 0.413 & 0.700 \\
			
			\multicolumn{1}{c|}{PG-SimCLR} & \underline{0.356} & \underline{0.837}  & \underline{0.361} & \underline{1.285} & \underline{0.275} & \underline{2.368} & \underline{0.269} &  \multicolumn{1}{c|}{\underline{3.140}} & \underline{0.307} & \underline{0.858} & \underline{0.166} & \underline{1.596} & -0.223 & 0.735 & \underline{0.622} & \underline{0.075} & \underline{0.434} & \underline{0.687} \\
			\midrule
			\multicolumn{1}{c|}{KnowCL} & \textbf{0.479} & \textbf{0.752} & \textbf{0.532} & \textbf{1.100} & \textbf{0.493} & \textbf{1.979} & \textbf{0.443} & \multicolumn{1}{c|}{\textbf{2.741}} & \textbf{0.424} & \textbf{0.783} & \textbf{0.325} & \textbf{1.436} & \textbf{0.153} & \textbf{0.612} & \textbf{0.658} & \textbf{0.042} & \textbf{0.536} & \textbf{0.622} \\
			\multicolumn{1}{c|}{Improv.} & 34.5\%  & 10.2\% &  47.3\% & 14.4\% & 79.3\% & 16.4\% & 64.7\% & \multicolumn{1}{c|}{12.7\%}  & 38.1\% & \multicolumn{1}{c}{8.7\%} & 95.8\% & 10.0\% & 7.0\% & 0.5\% & 5.8\% & 44.0\% & 23.5\% & 9.5\%   \\ \bottomrule
		\end{tabular}
	\end{table*}

	\begin{table*}[htbp]
		\caption{Street view imagery-based socioeconomic prediction results on three datasets. Best results are in bold and the best results are underlined. The last row shows relative improvement in percentage.}
		\vspace{-10px}
		\setlength\tabcolsep{2.2pt}
		\label{tab:sv_res}
		 \def\arraystretch{0.95}%
		\begin{tabular}{c|cccccccc|cccc|cccccc}
			\toprule
			\textbf{Dataset} & \multicolumn{8}{c|}{\textbf{Beijing}} & \multicolumn{4}{c|}{\textbf{Shanghai}} & \multicolumn{6}{c}{\textbf{New York}} \\
			\midrule
			\multirow{2}{*}{\textbf{Model}} & \multicolumn{2}{c}{\textbf{Pop.}} & \multicolumn{2}{c}{\specialcell{\textbf{Econ.}}} & \multicolumn{2}{c}{\specialcell{\textbf{Rest.}}} & \multicolumn{2}{c|}{{\textbf{Consp.}}} & \multicolumn{2}{c}{\textbf{Pop.}} & \multicolumn{2}{c|}{{\specialcell{\textbf{Econ.}}}} & \multicolumn{2}{c}{\textbf{Pop.}} & \multicolumn{2}{c}{\textbf{Edu.}} & \multicolumn{2}{c}{\textbf{Crime}} \\ 
			&  $R^2$ & RMSE & $R^2$ & RMSE & $R^2$ & RMSE & $R^2$ & RMSE & $R^2$ & RMSE & $R^2$ & RMSE & $R^2$ & RMSE & $R^2$ & RMSE & $R^2$ & RMSE \\
			\midrule
			\multicolumn{1}{c|}{ResNet-18} & 0.085 & 0.997  & 0.215  & 1.423 &  0.262 & 2.388 &  {0.257} & \multicolumn{1}{c|}{3.166} & 0.046 & 1.007 & {0.033} & \multicolumn{1}{c|}{1.718} & 0.151 & 0.612 & 0.402  & 0.095 &  0.340 & 0.742 \\
			
			\multicolumn{1}{c|}{Urban2Vec} & 0.026  & 1.029 & 0.059 & 1.559 & 0.094  & 2.646 & 0.103 & \multicolumn{1}{c|}{3.478} &  0.012 & 1.025 & 0.007 & 1.741 & 0.046 & 0.649 & 0.232 & 0.107  & 0.020  & 0.904  \\
			
			\multicolumn{1}{c|}{SceneParse} & 0.073 & 1.004 & 0.157 & 1.476 & 0.183 & 2.512 & {0.193}  &  \multicolumn{1}{c|}{3.299} & \underline{0.058} & \underline{1.001} & {0.049} & 1.704 & 0.154 & 0.611 & 0.426 & 0.093 & 0.222 & 0.806 \\
			
			\multicolumn{1}{c|}{PG-SimCLR} & \underline{0.237} & \underline{0.911}  & \underline{0.288} & \underline{1.356} & \underline{0.409} & \underline{2.136} & \underline{0.439} &  \multicolumn{1}{c|}{\underline{2.750}} & {0.015} & {1.023} & \underline{0.112} & \underline{1.646} & \underline{0.283} & \underline{0.563} & \underline{0.569} & \underline{0.080} & \underline{0.482} & \underline{0.657} \\
			\midrule
			\multicolumn{1}{c|}{KnowCL} & \textbf{0.416} & \textbf{0.796} & \textbf{0.557} & \textbf{1.069} & \textbf{0.470} & \textbf{2.024} & \textbf{0.449} & \multicolumn{1}{c|}{\textbf{2.725}} & \textbf{0.359} & \textbf{0.826} & \textbf{0.281} & \textbf{1.482} & \textbf{0.377} & \textbf{0.524} & \textbf{0.586} & \textbf{0.079} & \textbf{0.552} & \textbf{0.612} \\
			\multicolumn{1}{c|}{Improv.} & 75.6\%  & 12.6\% &  44.3\% & 21.2\% & 14.9\% & 5.2\% & 2.3\% & \multicolumn{1}{c|}{1.0\%}  & 519.0\% & \multicolumn{1}{c}{17.5\%} & 150.9\% & 10.0\% &  33.2\% & 6.9\% & 3.0\% & 1.3\% & 14.5\% & 6.8\%   \\ \bottomrule
		\end{tabular}
	\end{table*}

	\subsubsection{Baselines}
	We compare our model with several baselines in urban imagery-based socioeconomic prediction studies. The satellite imagery-based baselines include:
	\begin{itemize}[leftmargin=10px]
		\item \textbf{Tile2Vec} \citep{jean2019tile2vec}. Tile2Vec uses the triplet loss to minimize visual representations of spatially near satellite images and maximize those of distant pairs.
		\item \textbf{READ} \citep{han2020lightweight}. READ uses limited label data to train a teacher-student network with satellite imagery. The pre-trained model in original paper is used for comparison.
	\end{itemize}
	The street view imagery-based baselines includes:
	\begin{itemize}[leftmargin=10px]
		\item \textbf{Urban2Vec} \citep{wang2020urban2vec}. Urban2Vec follows the similar design with Tile2Vec but focuses on street view imagery.
		\item \textbf{SceneParse} \citep{zhou2017scene}. We use this scene parsing model to extract street view imagery representations following \citep{li2022predicting,lee2021predicting}.
	\end{itemize}
	We also apply two baselines for both satellite and street view imagery-based socioeconomic prediction:
	\begin{itemize}[leftmargin=10px]
		\item \textbf{ResNet-18} \citep{he2016deep}. ResNet-18 is pre-trained on ImageNet \citep{deng2009imagenet}, which is a backbone adopted in several related studies, and thus selected for comparison in both satellite and street view imagery.
		\item \textbf{PG-SimCLR} \citep{xi2022beyond}. PG-SimCLR originally employs SimCLR \citep{chen2020simple} for satellite imagery-based socioeconomic prediction, with spatiality and POI category distribution considered in similarity metric design. We select it for both satellite and street view imagery cases considering its competitive performance. 
	\end{itemize}
	
	We implement the baselines following reported settings or using pre-trained models in their original papers, and the obtained urban imagery representations are fed into the MLP-based socioeconomic indicator regression module for training and prediction. 
	
	\subsubsection{Metrics \& Implementation}
	We adopt the widely used rooted mean squared error (RMSE) and coefficient of determination ($R^2$) \citep{xi2022beyond,jean2016combining,han2020lightweight} for evaluation metrics. For the implementation, ResNet-18 \citep{he2016deep} and CompGCN \citep{vashishth2019composition} are adopted for visual and semantic encoders, respectively. We select Adam optimizer for parameter learning. In the contrastive learning step, we set the KG embedding dimension as 64 while the number of GCN layers is selected from $\{1,2,3,4\}$. The learning rate is set as 0.0003. In the socioeconomic prediction step, for each region, the learnt single urban imagery representation vector is used to predict various socioeconomic indicators with the learning rate and dropout searched from $\{0.0005, 0.001, 0.005\}$ and $\{0.1, 0.3, 0.5\}$. Besides, we randomly select 10 street view images for each region in the main experiment. The implementation codes are available at the link\footnote{\url{https://github.com/tsinghua-fib-lab/UrbanKG-KnowCL}}.

	\subsection{Performance Comparison}

	We evaluate the satellite imagery-based socioeconomic prediction performance in Table~\ref{tab:si_res}. Results on three datasets across six types of socioeconomic indicators demonstrate the superiority of our proposed KnowCL model, which improves the best baseline (PG-SimCLR) by 7\%-79\% on $R^2$ in all cases. Especially, KnowCL achieves the significant performance improvements owing to the comprehensive knowledge infused by the cross-modality based contrastive learning. As for the performance comparison with contrastive learning based models like Tile2Vec and PG-SimCLR, the results show that introducing more knowledge can bring better performance. For example, Tile2Vec only considers spatiality knowledge in similarity metric design while PG-SimCLR further considers function knowledge, leading to over 15\% improvements on $R^2$ on average. Besides, traditional models focus on satellite images in grid shape, failing to extract high-quality visual representations for the more practical case of irregular shape partitioned by road network.  
	
	As for the street view imagery-based socioeconomic prediction performance in Table~\ref{tab:sv_res}, KnowCL also achieves state-of-the-art results, which further validate the effectiveness and robustness. Limited by the repetitive street view images collected in Shanghai dataset, all baselines perform poorly across population and economic activeness prediction tasks therein, while KnowCL leverages UrbanKGs for informative representation learning from urban imagery with competitive performance achieved. 
	
	According to the absolute performance in Table~\ref{tab:si_res} and Table~\ref{tab:sv_res}, the socioeconomic indicators of different cities show diverse preference to urban imagery, e.g., KnowCL with satellite imagery obtains the best absolute performance for population prediction in Beijing and Shanghai, while the street view imagery becomes the better choice for population and crime prediction in New York. This phenomenon is mainly determined by city structures and socioeconomic indicator characteristics. Different from complex city structures in Beijing and Shanghai, New York follows the grid layout with block regions in similar shapes, which provides limited information for population estimation. Besides, the street view imagery can provide an internal view for urban environment safety perception, as validated in \citep{naik2014streetscore}. Such results further indicate that both satellite and street view imagery can provide complementary information to each other, and our proposed KnowCL model can fully exploit the value of urban imagery, which is quite essential for urban environment perception and SDG monitoring.

	\begin{figure}[hbtp]
		\vspace{-10px}
		\centering
		\subfigure[Satellite Imagery]{
			\includegraphics[width=0.2\textwidth]{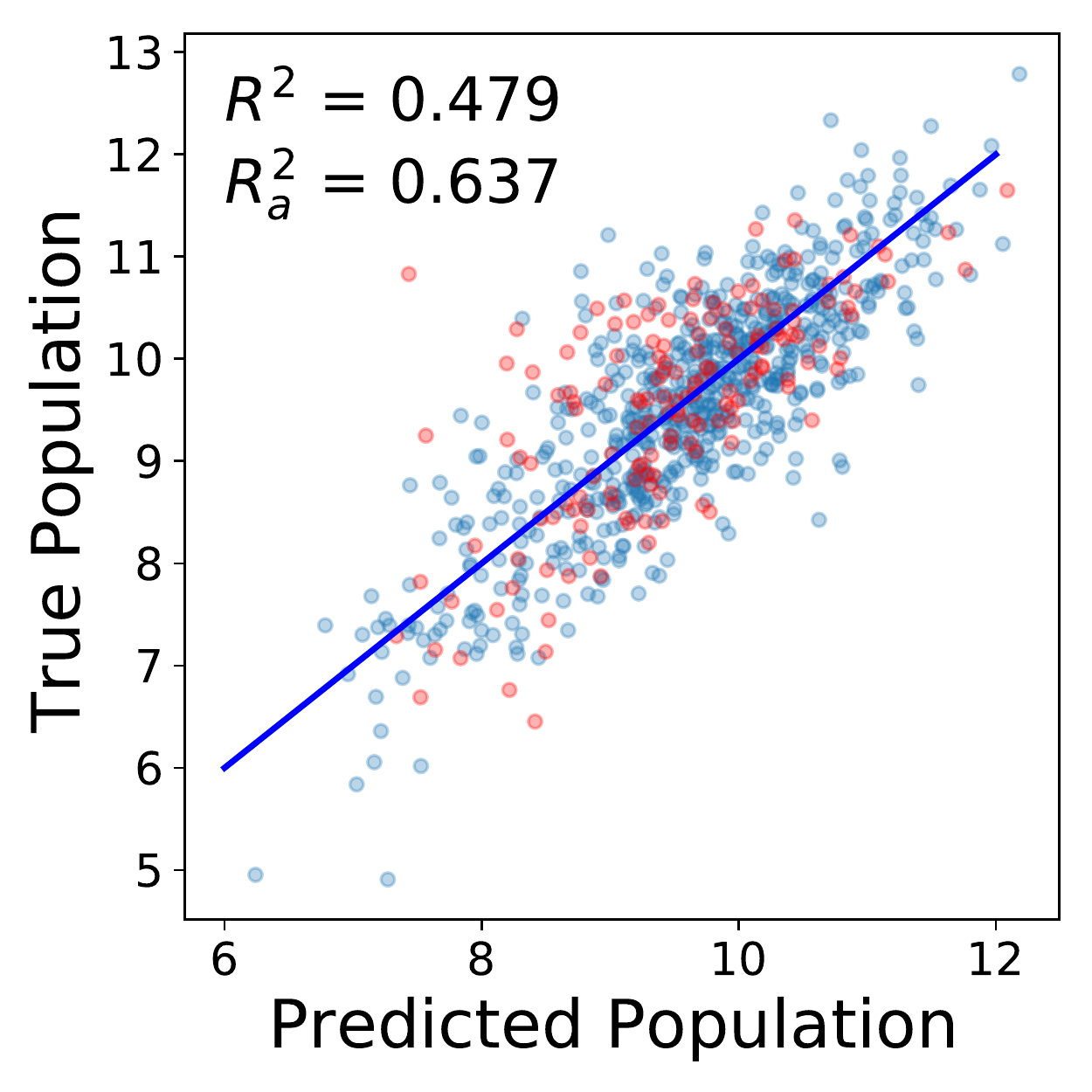}}
		\subfigure[Street View Imagery]{
			\includegraphics[width=0.2\textwidth]{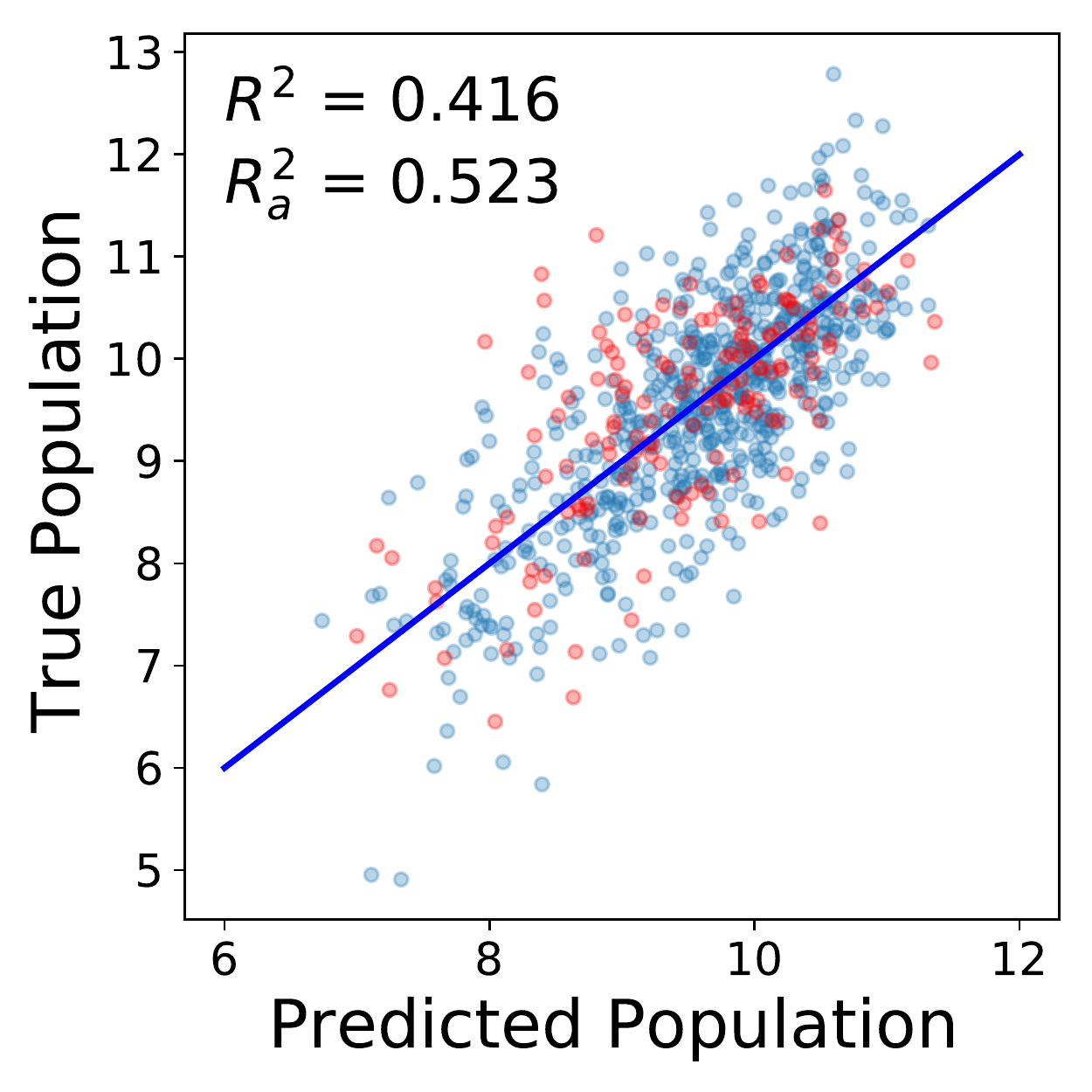}}
		\vspace{-10px}
		\caption{Predicted population versus true population across all regions on Beijing dataset. Blue line is at 45\textdegree. $R^2$ and $R^2_a$ correspond to the results of testing regions (red dots) and all regions (red and blue dots), respectively}\label{fig:r2}
		\vspace{-5px}
	\end{figure}
	
	To further analyze the predictive power of our proposed KnowCL model, in Figure~\ref{fig:r2}, we compare the predicted and true population for all regions in Beijing dataset, and results for other datasets are provided in Section~\ref{sec:add_exp}. The results show that KnowCL can well replicate the population of most regions (see the dots along 45\textdegree\ line) via urban imagery, explaining 52\%-63\% of the variation in population on two datasets.

	\subsection{Ablation Study}
	\subsubsection{Effectiveness of Knowledge Identification}
\begin{figure}[hbtp]
		\vspace{-10px}
		\centering
		\subfigure[Satellite Imagery]{
			\includegraphics[width=0.23\textwidth]{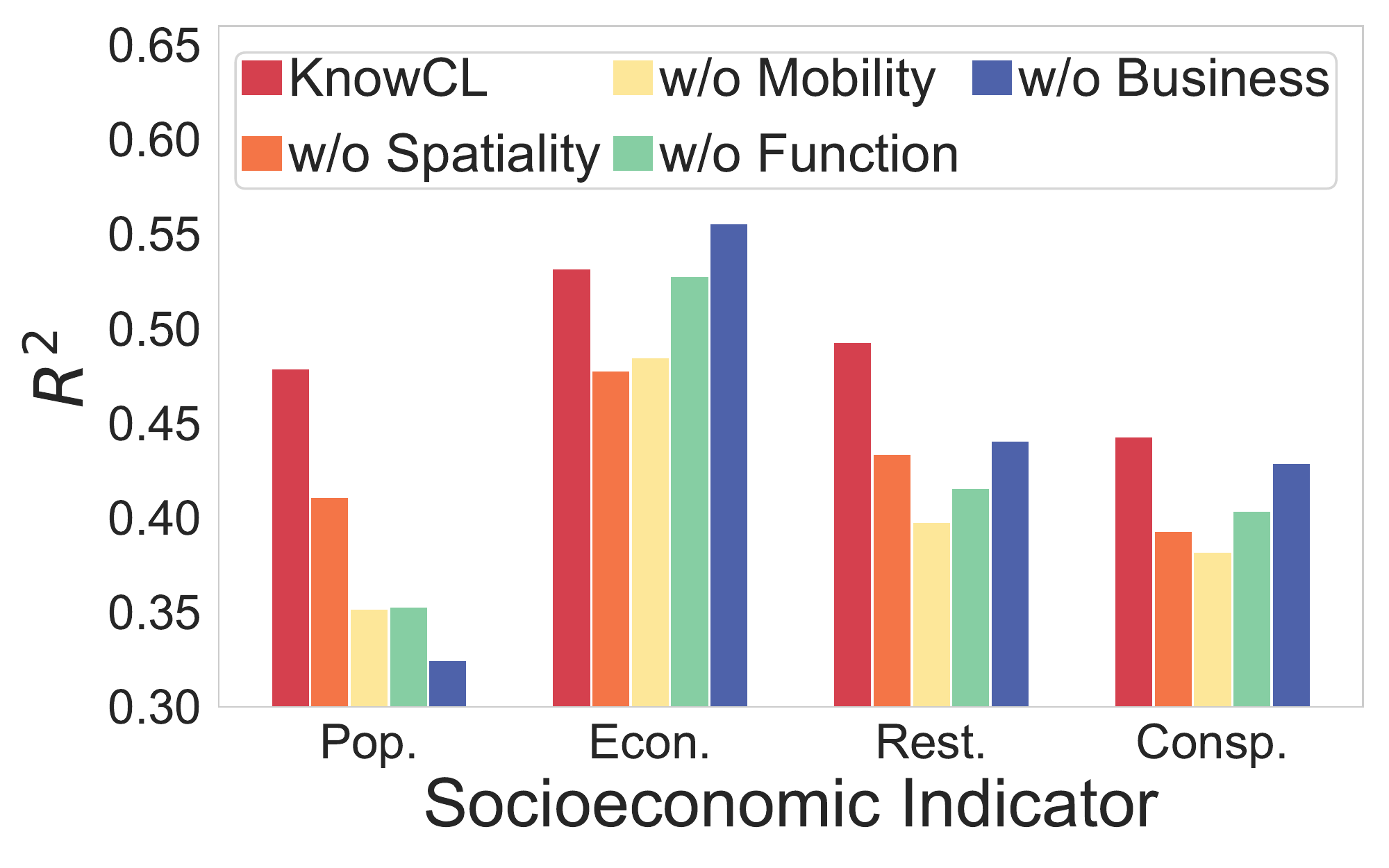}}
		\subfigure[Street View Imagery]{
			\includegraphics[width=0.23\textwidth]{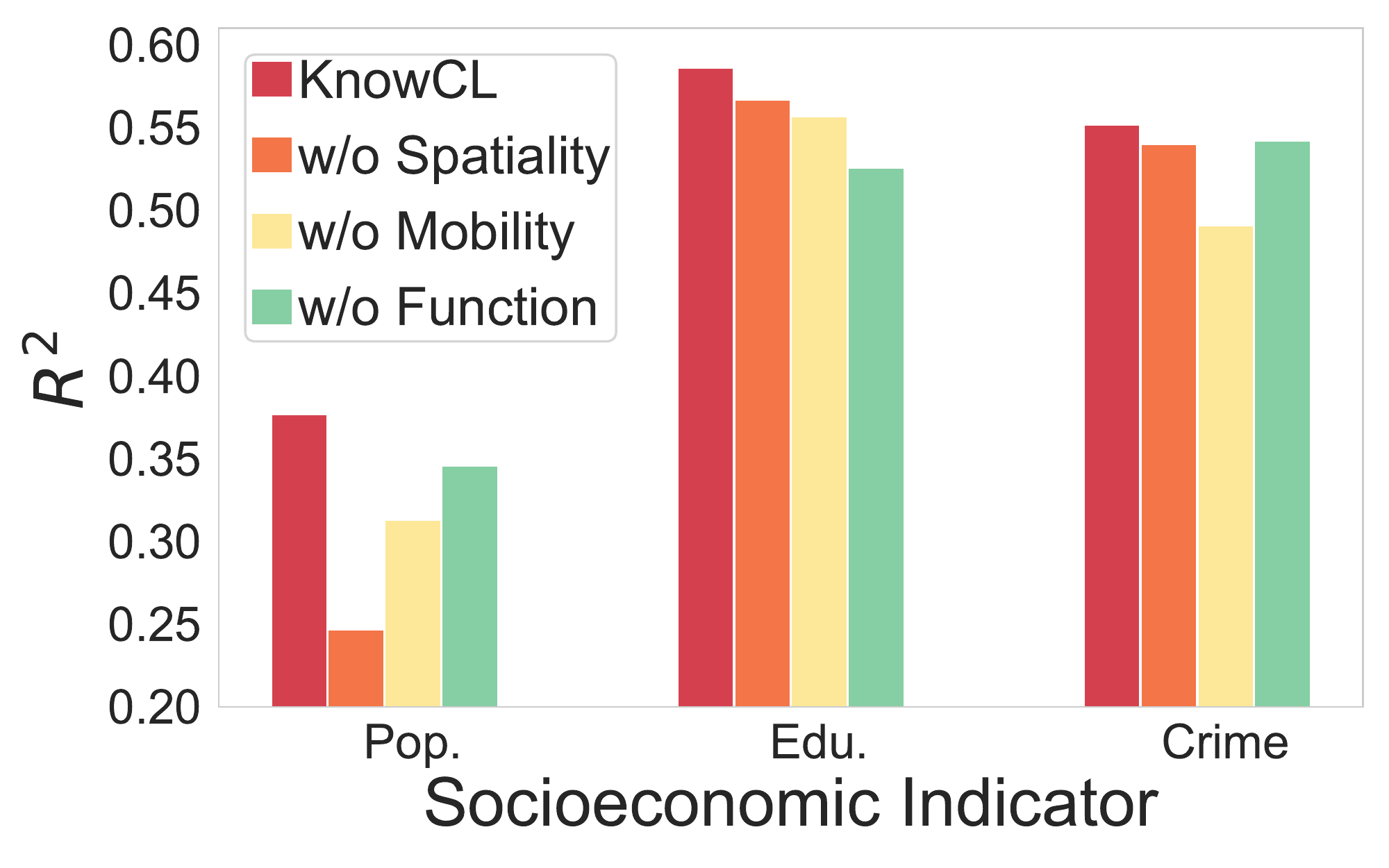}}
		\vspace{-10px}
		\caption{Performance comparison of different identified knowledge on Beijing and New York datasets with satellite and street view imagery, respectively.}\label{fig:abl_kg}
				\vspace{-10px}
	\end{figure}

	To validate the effectiveness of identified knowledge in UrbanKG, Figure~\ref{fig:abl_kg} presents the performance comparison of UrbanKG without certain type of knowledge. We select Beijing and New York datasets for evaluation, on which satellite and street view imagery inputs achieve the best absolute performance, respectively. The business knowledge on New York dataset is not provided in UrbanKG and thus not reported.
	
	All four types of semantic knowledge identified by UrbanKG are essential for socioeconomic prediction, according to the findings in Figure~\ref{fig:abl_kg}. Particularly, the knowledge that is hardly captured by urban imagery is more important, e.g., the mobility knowledge of crowd flow transitions between regions brings 5\%-30\% gains for predicting all socioeconomic indicators, because both satellite and street view imagery cannot capture such dynamic information without additional knowledge infused. Additionally, the impacts of various types of semantic knowledge vary to socioeconomic indicators. For example, in New York dataset, the education indicator is highly correlated with function knowledge while the population indicator prefers to spatiality knowledge, which enlightens us to identify comprehensive knowledge for a broader urban imagery-based socioeconomic prediction with more indicators considered.

	\subsubsection{Effectiveness of Knowledge Infusion}
	A major novelty of this paper is introducing the cross-modality based contrastive learning with the image-KG contrastive loss for knowledge infusion, which is different from the single-modality based ones in existing studies \cite{jean2019tile2vec,xi2022beyond}. To validate the effectiveness, we develop a direct image-image contrastive loss for knowledge infusion, which is similar to PG-SimCLR \citep{xi2022beyond}, termed as KG-SimCLR. Specifically, KG-SimCLR calculates KG embedding similarity for positive region pairs, and requires their associated urban images to be closer in visual representation space.

\begin{table}[htbp]
	\caption{Performance comparison $R^2$ of different knowledge infusion ways on Beijing and New York datasets.}
	\vspace{-10px}
	\setlength\tabcolsep{2.2pt}
	\label{tab:abl_loss}
	 \def\arraystretch{0.95}%
	\begin{tabular}{cc|cccc|ccc}
		\toprule
		\textbf{} & \textbf{} & \multicolumn{4}{c|}{\textbf{Beijing}} & \multicolumn{3}{c}{\textbf{New York}} \\ \midrule
		 \multicolumn{2}{c|}{\textbf{Model}} & \textbf{Pop.} & \textbf{Econ.} & \textbf{Rest.} & \textbf{Consp.} & \textbf{Pop.} & \textbf{Edu.} & \textbf{Crime} \\ \midrule
		\multicolumn{1}{c|}{\multirow{2}{*}{\textbf{SI}}} & KG-SimCLR & 0.272 & 0.197 & 0.192 & 0.175 & -0.302 & 0.555 & 0.341 \\
		\multicolumn{1}{c|}{} & KnowCL & 0.479 & 0.532 & 0.493 & 0.443 & 0.153 & 0.658 & 0.536 \\ \midrule
		\multicolumn{1}{c|}{\multirow{2}{*}{\textbf{SV}}} & KG-SimCLR & 0.209 & 0.229 & 0.299 & 0.329 & 0.053 & 0.382 & 0.294 \\
		\multicolumn{1}{c|}{} & KnowCL & 0.416 & 0.557 & 0.470 & 0.449 & 0.377 & 0.586 & 0.552 \\ \bottomrule
	\end{tabular}
	\vspace{-5px}
\end{table}

\begin{figure}[htbp]
	\centering
	\includegraphics[width=.99\linewidth]{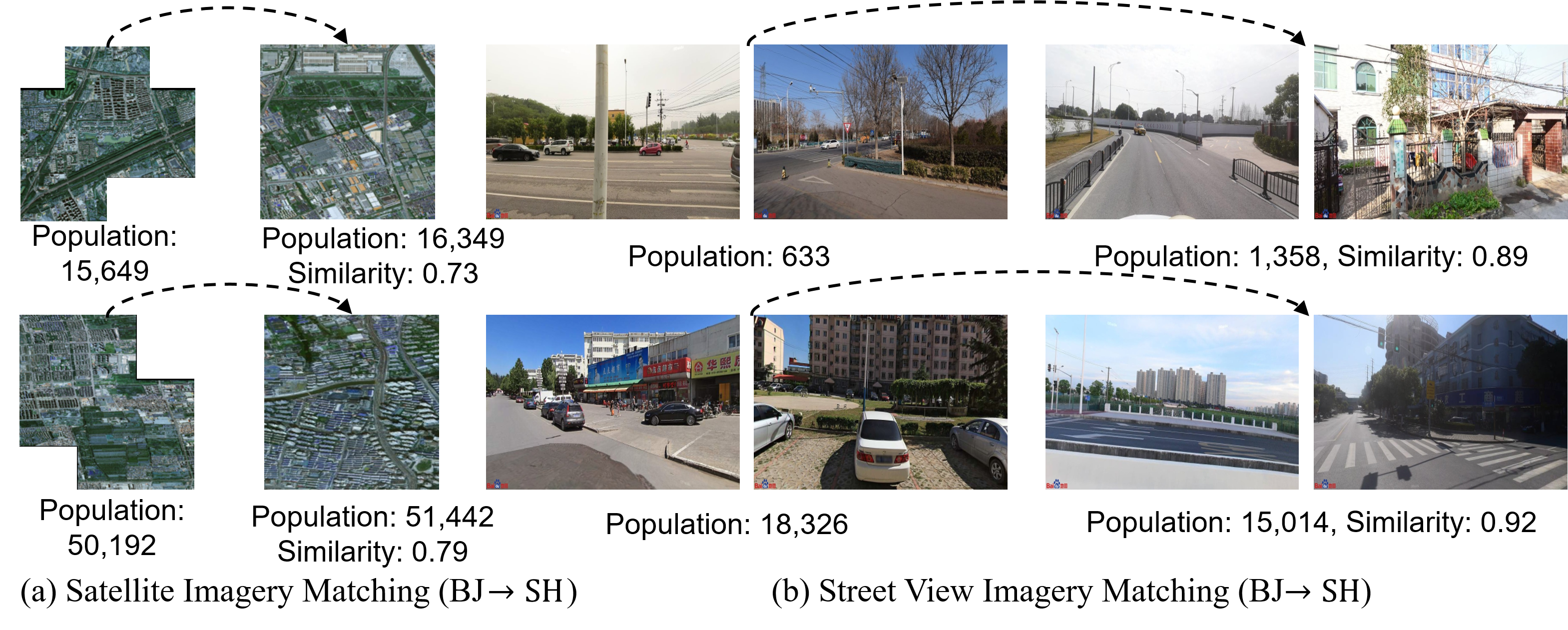}
	\vspace{-10px}
	\caption{Most similar urban imagery matching between Beijing and Shanghai datasets via learnt urban imagery representations by KnowCL. The population indicator and cosine similarity are presented below the images. Satellite images might be in irregular shape due to the shape of associated regions.}
	\label{fig:image_matching}
\end{figure}

\begin{figure}[htbp]
	\vspace{-10px}
	\centering
	\subfigure[Satellite Imagery]{
		\includegraphics[width=0.2\textwidth]{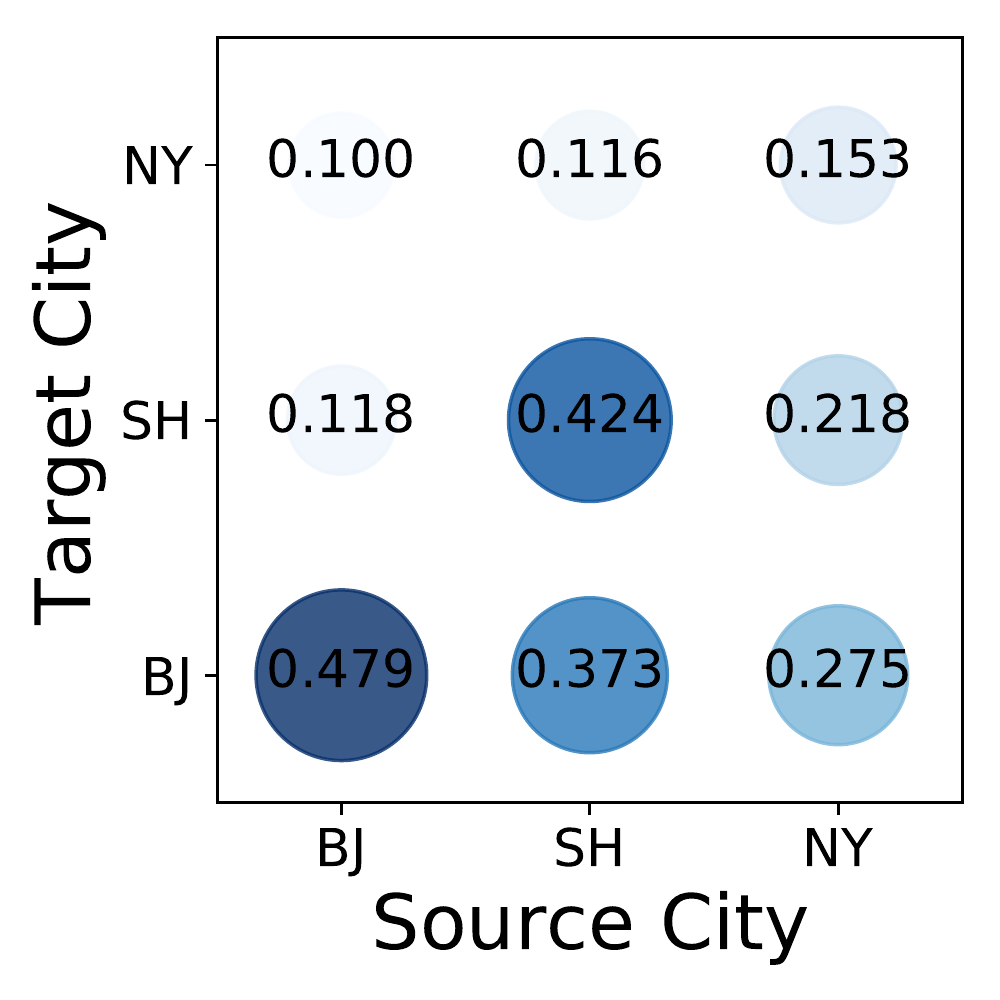}}
	\subfigure[Street View Imagery]{
		\includegraphics[width=0.2\textwidth]{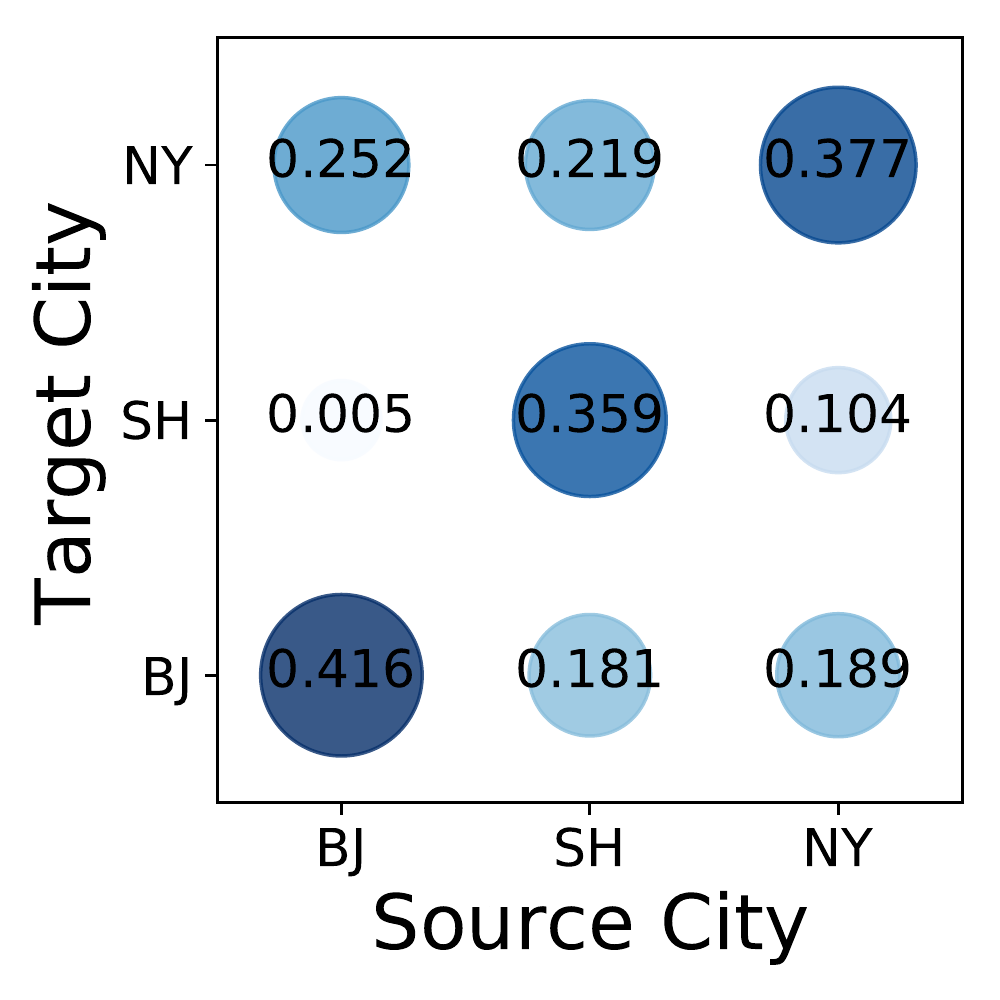}}
	\vspace{-10px}
	\caption{The $R^2$ for the transferability test on satellite and street view imagery-based population prediction.}\label{fig:transfer}
	\vspace{-10px}
\end{figure}

	Table~\ref{tab:abl_loss} presents the performance comparison between KG-SimCLR and KnowCL with different urban imagery inputs. The significant performance gaps between two models on both satellite and street view imagery-based socioeconomic prediction indicate that simply using existing image view-based contrastive loss cannot achieve effective knowledge infusion. Especially, directly modeling the comprehensive knowledge in KG as a similarity metric provides a quite weak self-supervision signal for visual representation learning, while our proposed cross-modality based contrastive learning framework infuses such knowledge via similarity matching in representation space. Overall, the ablation studies demonstrate the effectiveness of our proposed knowledge infusion design and can potentially apply in various urban imagery-based research.

	\subsection{Transferability Study}	
	\subsubsection{Prediction Performance Across Cities/Countries}
	The experiment results above validate the effectiveness of UrbanKG, which however might be not available in underdeveloped and developing cities/countries due to data deficiency. Thus, here we investigate the practical case of socioeconomic prediction in transfer setting \citep{park2022learning}: Given the visual encoder of a KnowCL model trained on a source city with urban imagery and UrbanKG data, we apply it for socioeconomic prediction in target cities where only urban imagery data are available. The transferability task checks whether KnowCL infuses shared knowledge across cities into the visual encoder for urban imagery-based socioeconomic prediction.

	We vary the source-target city pairs and report the population prediction performance in Figure~\ref{fig:transfer}, where both satellite imagery and street view imagery are evaluated. Here 40 street view images for each region in off-diagonal transfer experiments are used for robust performance. According to the results, the diagonal line shows the highest correlation due to the same city transferred from the source to the target. Moreover, compared with baselines trained and evaluated on the same dataset in Table~\ref{tab:si_res} and Table~\ref{tab:sv_res}, KnowCL achieves competitive transfer performance for both satellite and street view imagery-based population prediction in Bejing and New York datasets, as validated by similar scatter sizes in each row. For example, SH$\rightarrow$BJ transfer experiment achieves a $R^2$ of 0.373 compared with 0.356 of the best baseline (PG-SimCLR) achieved in non-transfer setting. Such results validate the transferability of our proposed KnowCL model for socioeconomic prediction across cities/countries, which mainly owes to the shared knowledge identified in UrbanKG and infused in visual encoder by cross-modality based contrastive learning. Hence, pre-trained KnowCL model can be leveraged for socioeconomic prediction in cities without UrbanKG. Besides, we also investigate the transferability of the best baseline PG-SimCLR in Section~\ref{sec:add_exp}, which is less competitive due to limited knowledge considered.
	
	\subsubsection{Visual Analogies Across Cities}
	We also investigate the visual similarity between urban imagery across cities. Specifically, given an urban image in source city, we compute the cosine similarity between its visual representation and all visual representations in another city, and select the most similar ones for comparison \citep{wang2020urban2vec}, as shown in Figure~\ref{fig:image_matching}. As for the satellite imagery matching in Figure~\ref{fig:image_matching}(a), similar regions share the similar distribution of buildings as well as populations. On the other hand, the street view imagery matching in Figure~\ref{fig:image_matching}(b) successfully identifies regions with similar physical appearance and populations. Thus, the knowledge-infused urban imagery representations capture not only visual features but also socioeconomic information associated with regions.

	\section{Conclusion}
	In this paper, we present a novel approach to make predictions on population, economic activity, consumption, education, and public safety indicators from web-collected urban imagery covering both satellite and street view imagery. 
	Our proposed knowledge-infused contrastive learning model KnowCL is built upon the comprehensive knowledge identified by urban knowledge graph, and further designs an image-KG contrastive loss for effective knowledge infusion into urban imagery representations. KnowCL is the first solution that introduces the knowledge graph and the cross-modality based contrastive learning framework for urban imagery-based socioeconomic prediction.
	Extensive experiments validate the model effectiveness and transferability in different cities across several socioeconomic indicators. 
	
	We demonstrated model's potential in socioeconomic prediction with ubiquitous urban imagery, which is of great importance to sustainable development in data-poor regions and countries. Although our model outperforms baselines, the results may be less interpretable, so in future work we will consider exploring in-depth the semantics of UrbanKG for interpretability.

	\begin{acks}
		This work was supported in part by 	the National Key Research and Development Program of China under 2020AAA0106000, the National Nature Science Foundation of China under 61972223, 61971267, U1936217. 
	\end{acks}

		\bibliography{sample-base}
\bibliographystyle{ACM-Reference-Format}

	\newpage
	\appendix
	
	\section{Details of Dataset} \label{sec:add_data}
	\subsection{UrbanKG Construction}
\begin{figure}[hbtp]
	 	\centering
	 	\subfigure[Beijing]{
	 		\includegraphics[width=0.2\textwidth]{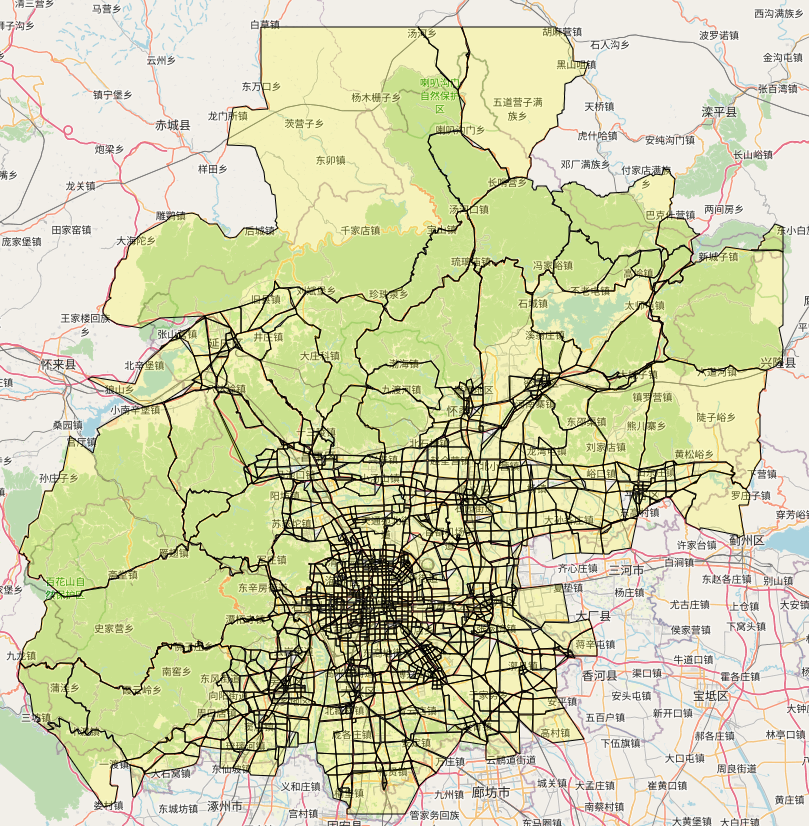}}
	 	\subfigure[Shanghai]{
	 		\includegraphics[width=0.2\textwidth]{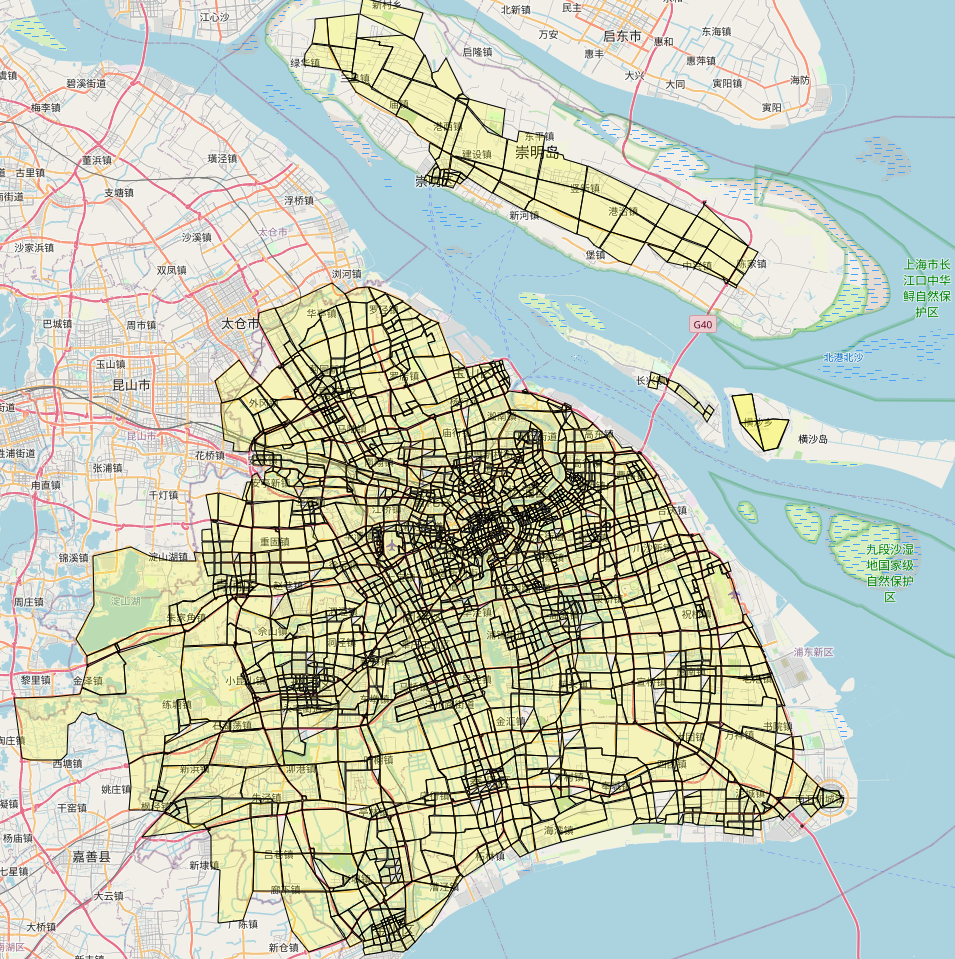}}
	 	\vspace{-10px}
	 	\caption{Visualization of region entities in UrbanKG for Beijing and Shanghai datasets.}\label{fig:regions}
	 \end{figure}

	Here we introduce the details of urbanKG construction. For region entities in UrbanKG for Beijing and Shanghai datasets, we partition the city into multiple regions by the road network, which are shown in Figure~\ref{fig:regions}. The region entities in New York dataset follows the CBG division by US Census Bureau, whose visualization can be referred to the official link\footnote{\url{https://data.cityofnewyork.us/City-Government/2010-Census-Blocks/v2h8-6mxf}}. Each region entity is provided with a sequence of longitude-latitude pairs $L_a=\{(lng_a^1,lat_a^1),\cdots,(lng_a^k, lat_a^k)\}$ as region boundary. POI entities and business center entities are provided with location information of longitude-latitude pairs like $l_i=(lng_i, lat_i)$. The category entities are POI properties identified by experts, e.g., food, shopping, accommodation, business, residence, education, etc. 
	 
	Based on the entities above, the relational links defined in Table~\ref{tab:relations} can be extracted as follows.
	\begin{itemize}[leftmargin=10px]
		\item \emph{borderBy.} Given two regions $a, b$, they are connected by \emph{borderBy} if $\vert L_a\cap L_b\vert>0$, i.e., sharing the same boundary points. 

		\item \emph{nearBy.} Given two regions $a, b$, they are connected by \emph{nearBy} if $\|\bar{L}_a-\bar{L}_b\|\leq 1km$, where $\bar{L}_a, \bar{L}_b$ are center location of regions.
		
		\item \emph{locateAt.} Given a POI $p$ and a region $a$, they are connected by \emph{locateAt} if $l_p$ is in the closure by region boundary $L_a$.
		
		\item \emph{flowTransition.} Given two regions $a,b$, they are connected by \emph{flowTransition} if the aggregated flow transition between two regions exceeds the threshold.
		
		\item \emph{similarFunction.} Given two regions $a,b$ and the category distribution vectors of POIs therein $\bm{z}_a, \bm{z}_b$, they are connected by \emph{similarrFunction} if $cos(\bm{z}_a,\bm{z}_b)\geq 0.95$ with cosine similarity.
		
		\item \emph{coCheckin.} Given two POIs $p_1, p_2$, they are connected by \emph{coCheckin} if the number of records that consecutively visit $p_1$ and $p_2$ exceeds the threshold. 
		
		\item \emph{cateOf.} A POI is connected to its associated category by \emph{cateOf}.
		
		\item \emph{provideService.} Given a region $a$ and a business center $bc$, they are connected by \emph{provideService} if $\|\bar{L}_a-l_{bc}\|\leq 3km$.
		
		\item \emph{belongTo.} Given a POI $p$ and a business center $bc$, they are connected by \emph{belongTo} if $\|l_p-l_{bc}\|\leq 3km$.
		
		\item \emph{competitive.} Given two POIs $p_1, p_2$, they are connected by \emph{competitive}  if $\|l_{p_1}-l_{p_2}\|\leq 500m$ and they are in the same category.
	\end{itemize}

\begin{figure*}[hbtp]
	\vspace{-10px}
	\centering
	\subfigure[Satellite Imagery (SH)]{
		\includegraphics[width=0.2\textwidth]{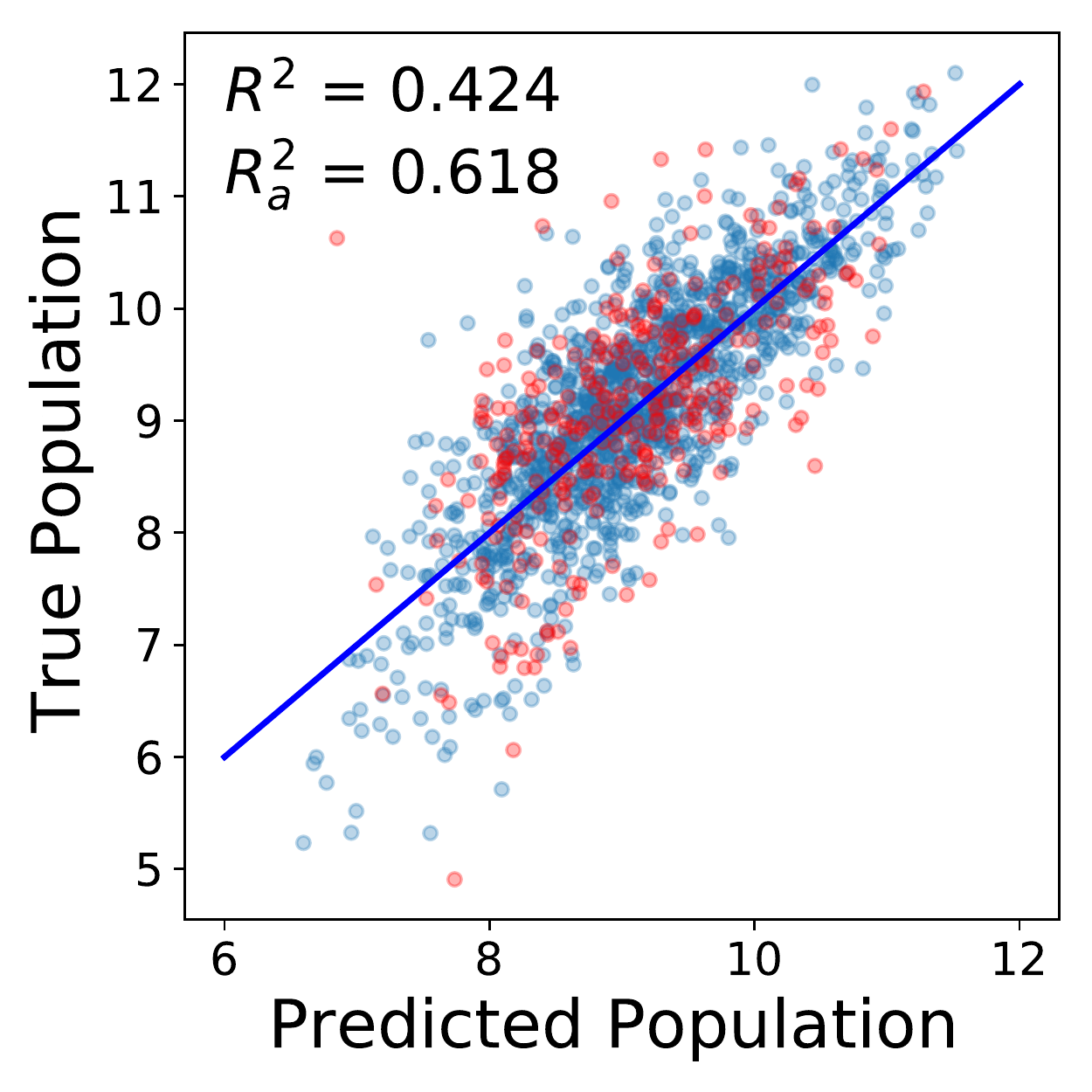}}
	\subfigure[Street View Imagery (SH)]{
		\includegraphics[width=0.2\textwidth]{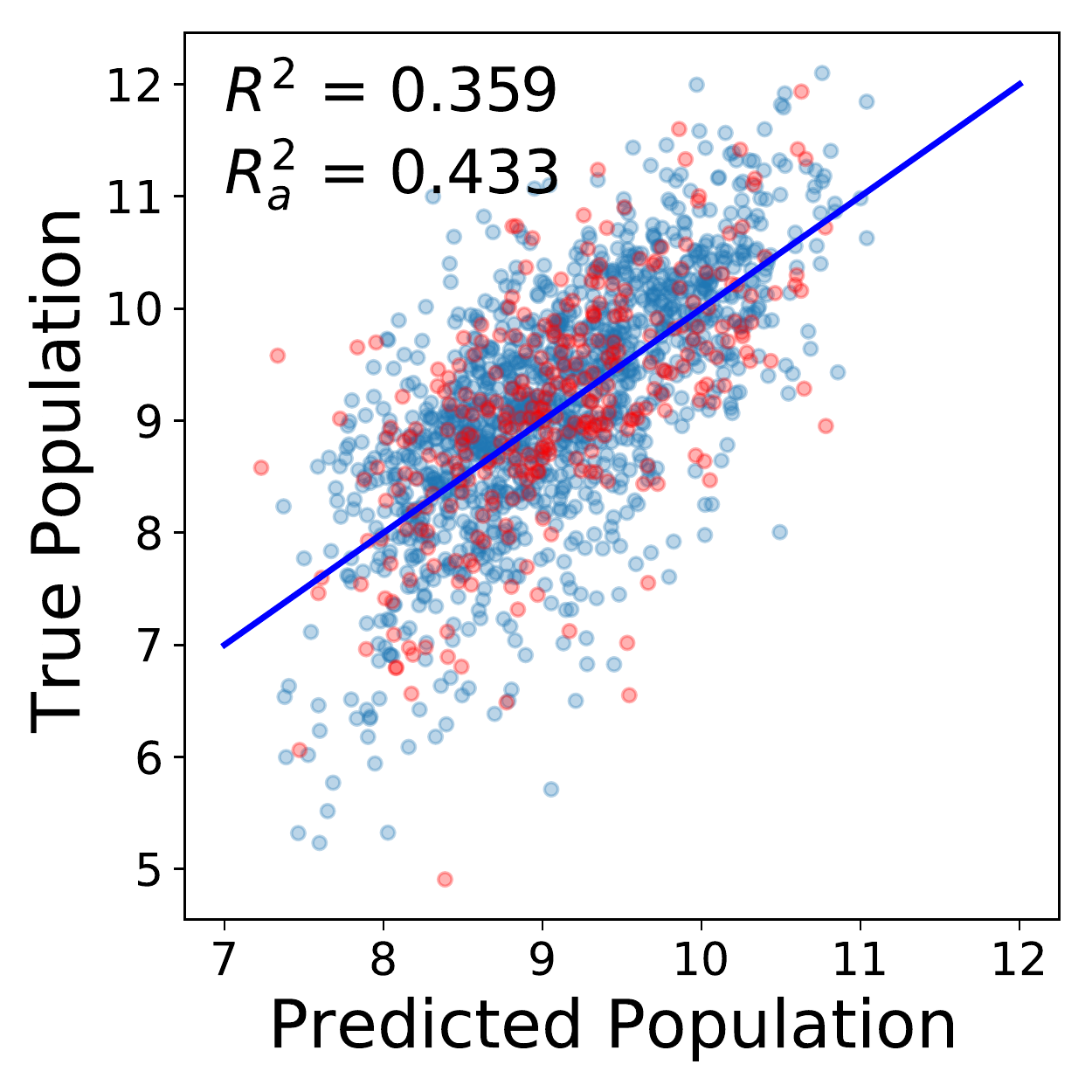}}
	\subfigure[Satellite Imagery (NY)]{
		\includegraphics[width=0.2\textwidth]{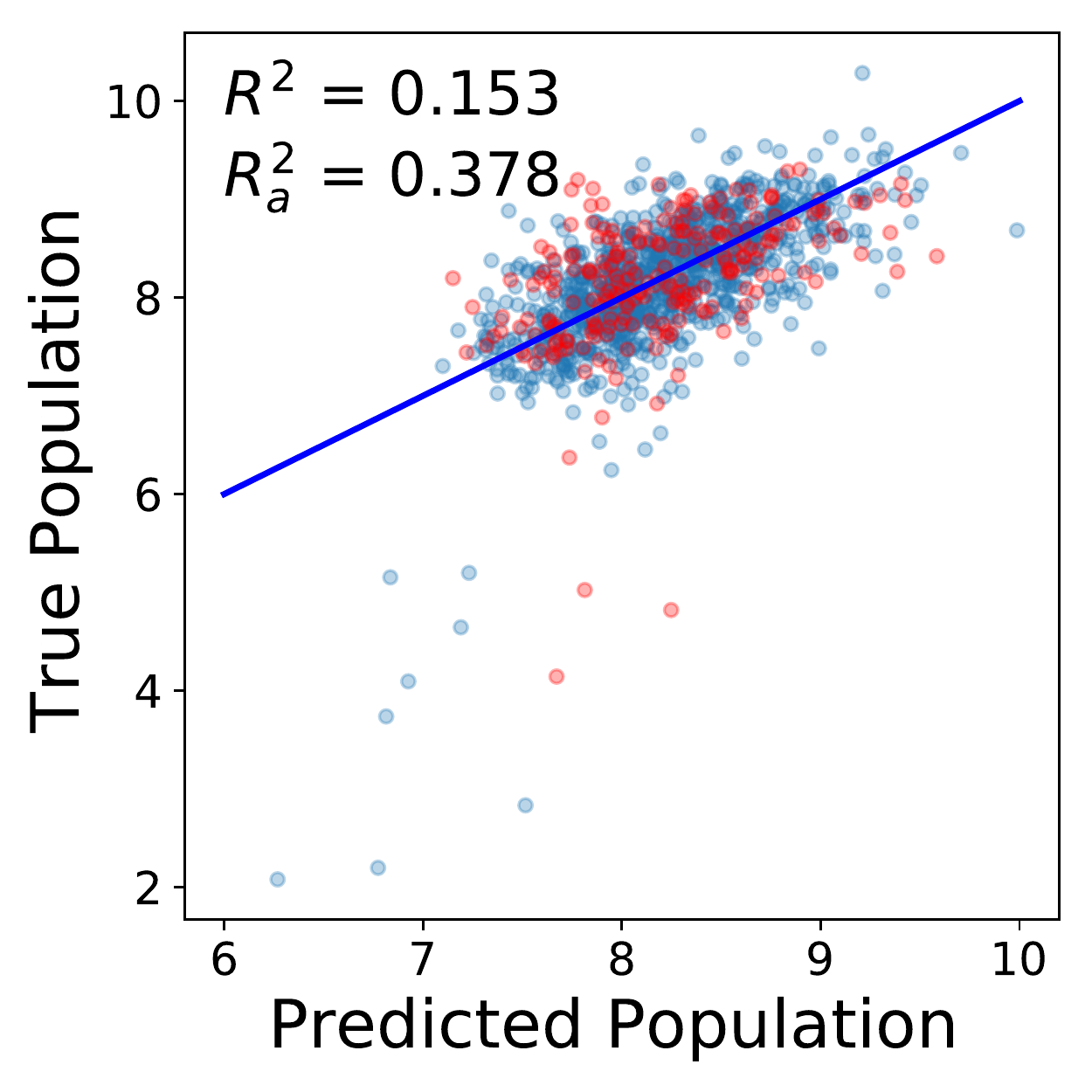}}
	\subfigure[Street View Imagery (NY)]{
		\includegraphics[width=0.2\textwidth]{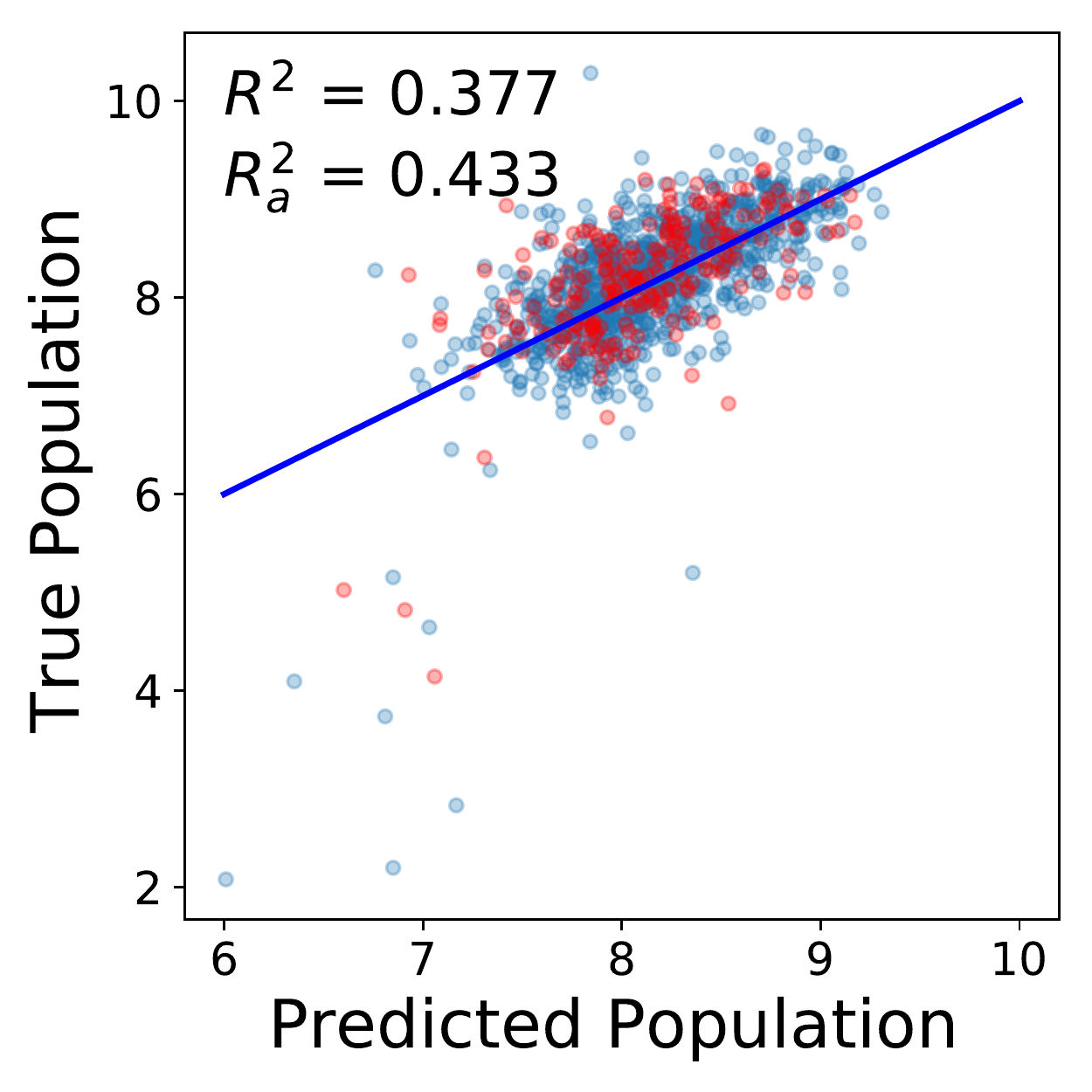}}
	\vspace{-12px}
	\caption{Predicted population versus true population across all regions on Shanghai and New York datasets. Blue line is at 45\textdegree. $R^2$ and $R^2_a$ correspond the results of testing regions (red dots) and all regions (red and blue dots), respectively}\label{fig:r2_sh_ny}
			\vspace{-5px}
\end{figure*}

\begin{figure*}[hbtp]
	\centering
	\subfigure[Satellite (BJ)]{
		\includegraphics[width=0.15\textwidth]{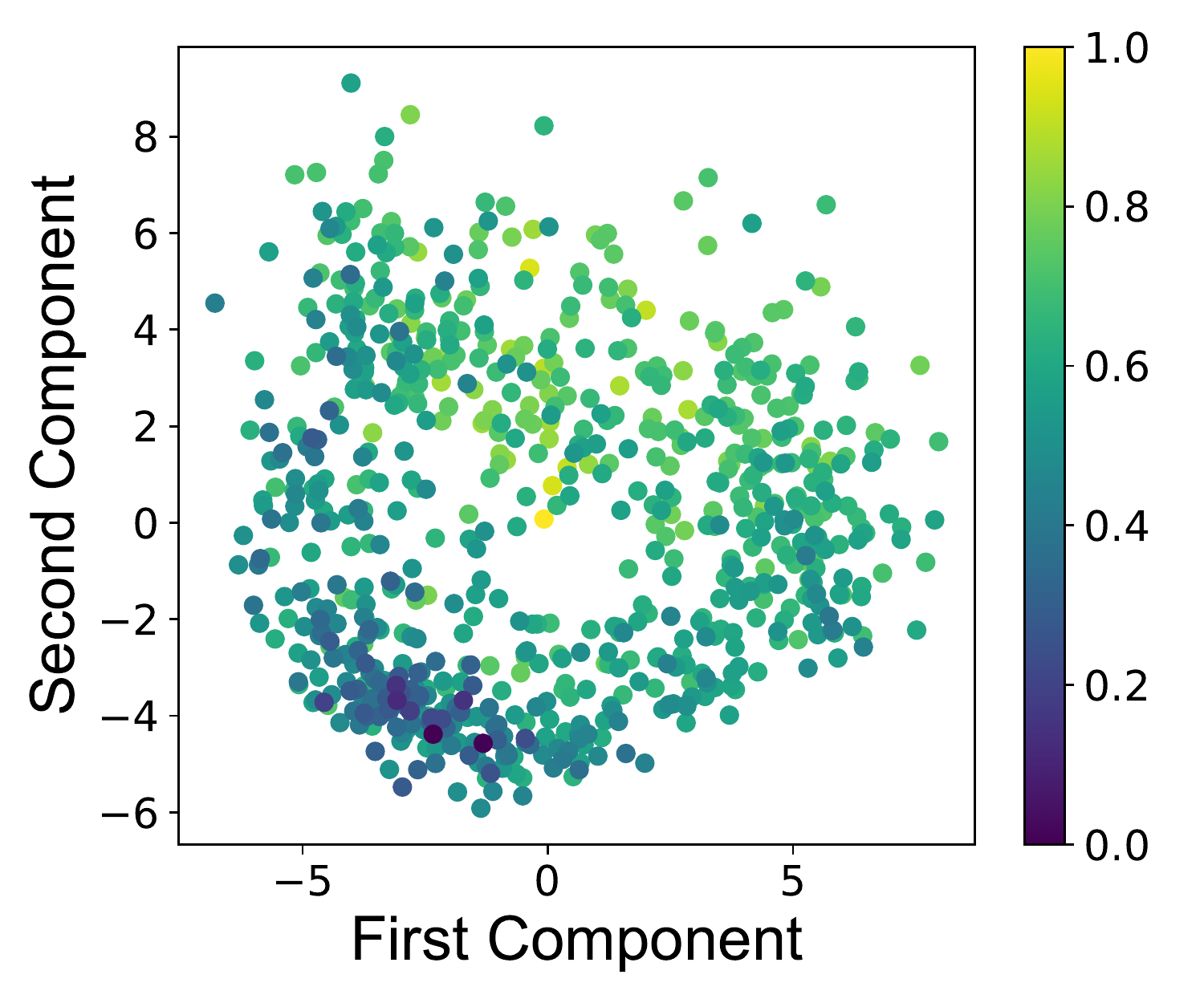}}
	\subfigure[Street View (BJ)]{
		\includegraphics[width=0.15\textwidth]{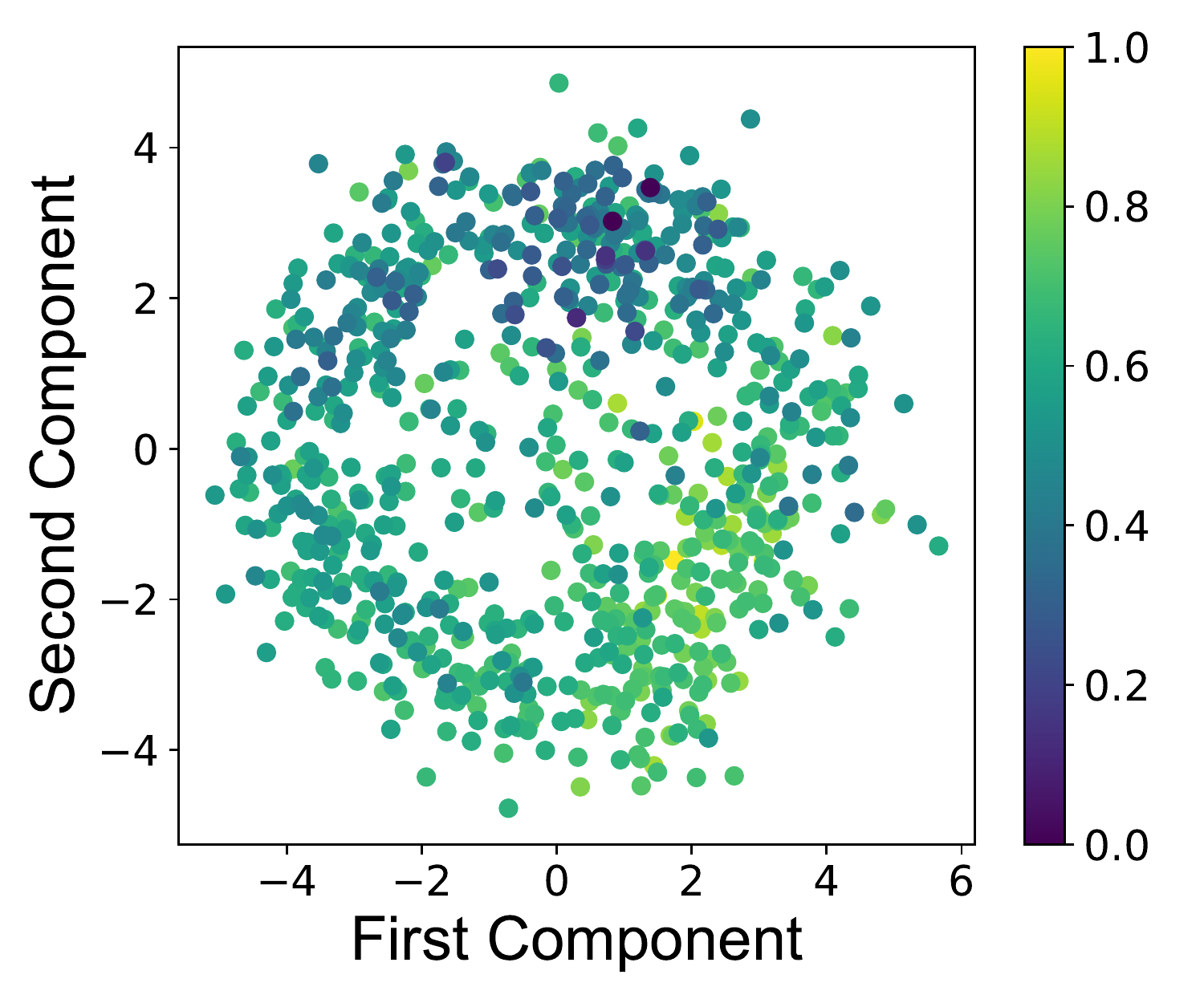}}
	\subfigure[Satellite (SH)]{
		\includegraphics[width=0.15\textwidth]{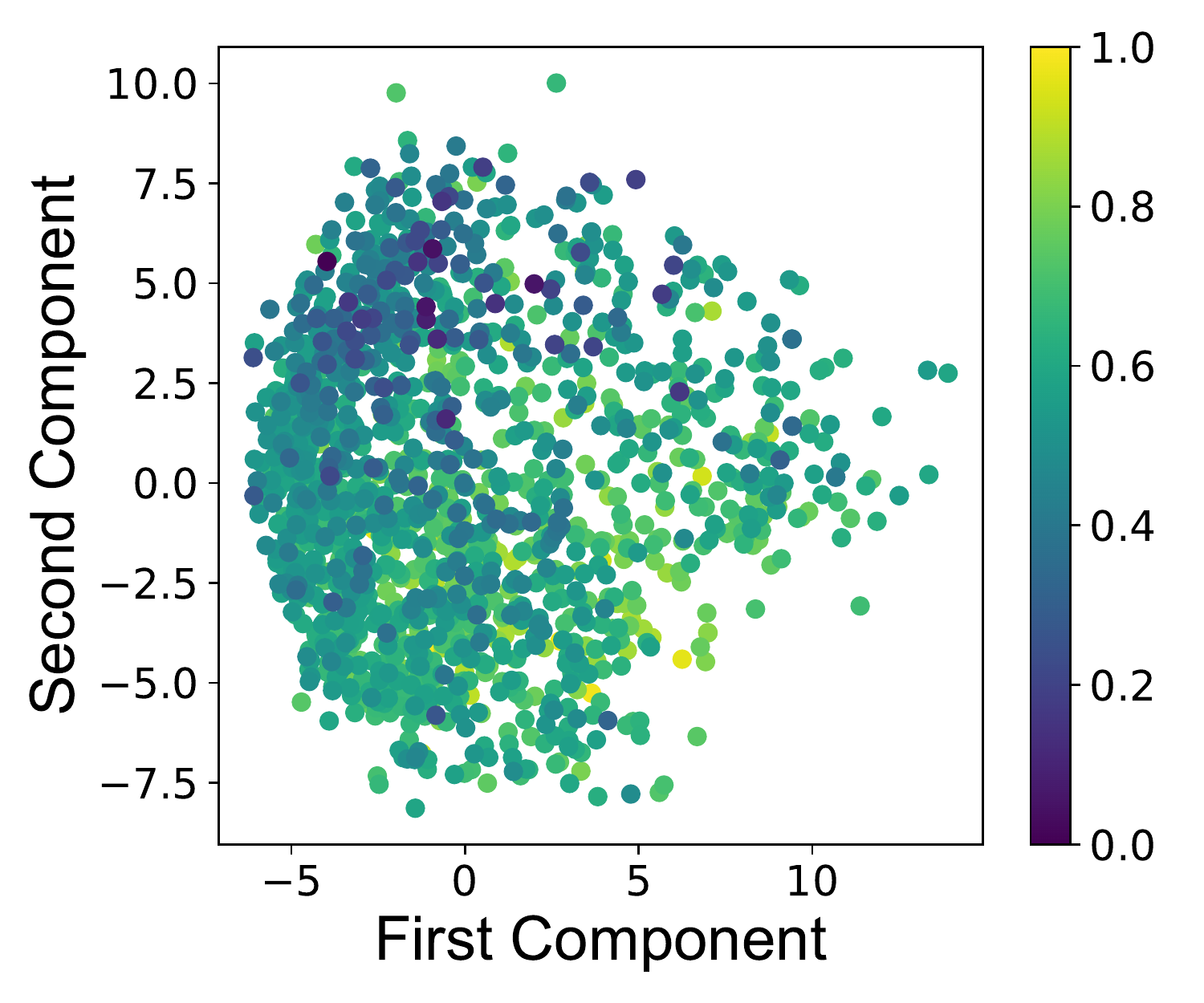}}
	\subfigure[Street View (SH)]{
		\includegraphics[width=0.15\textwidth]{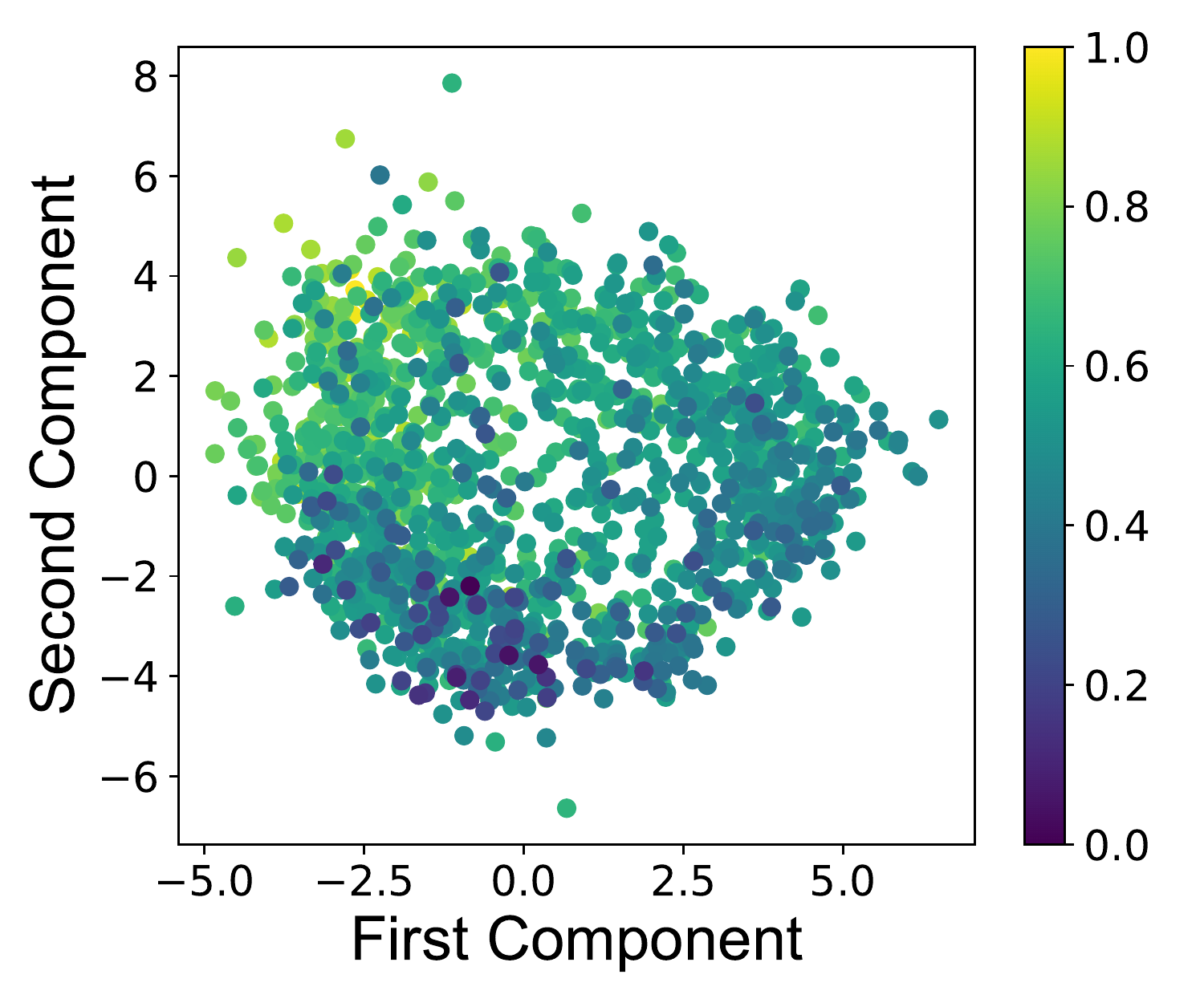}}
	\subfigure[Satellite (NY)]{
		\includegraphics[width=0.15\textwidth]{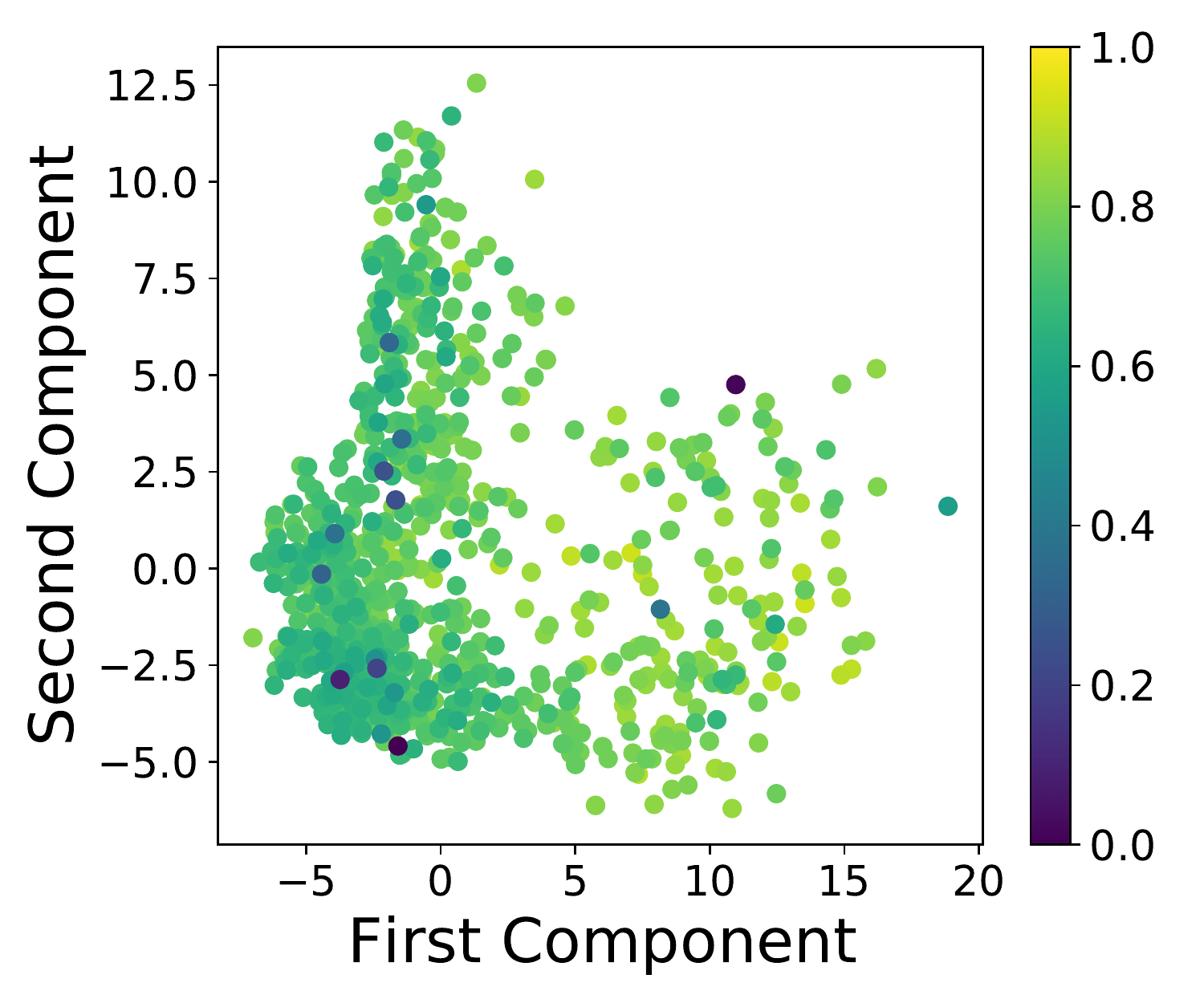}}
	\subfigure[Street View (NY)]{
		\includegraphics[width=0.15\textwidth]{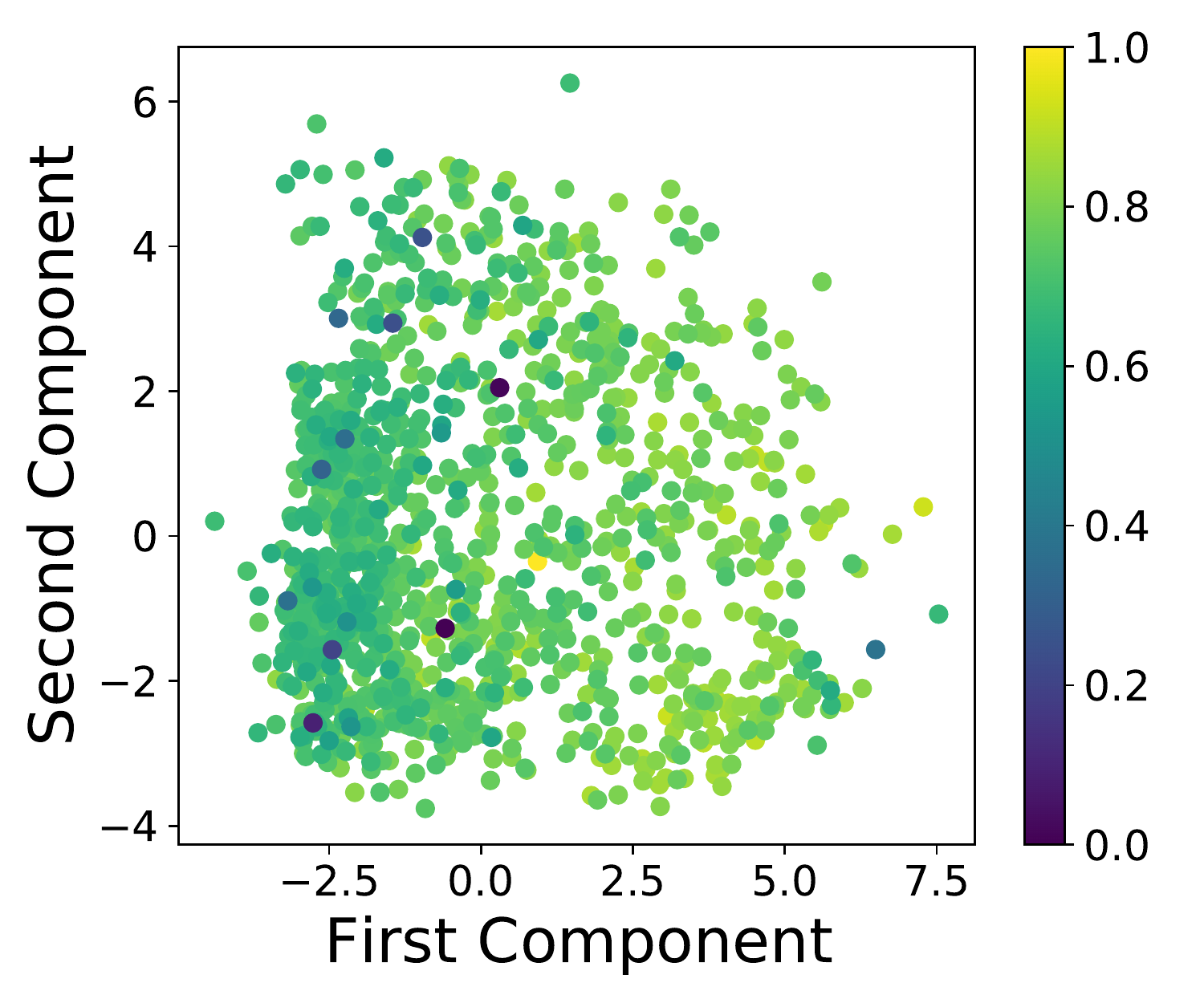}}
	\vspace{-10px}
	\caption{Visualization of urban imagery based on PCA algorithm, where dot color represents the value of corresponding population to the urban imagery.}
	\label{fig:pca_vis}
		\vspace{-10px}
\end{figure*}

\begin{figure}[hbtp]
	\centering
	\subfigure[Satellite Imagery]{
		\includegraphics[width=0.2\textwidth]{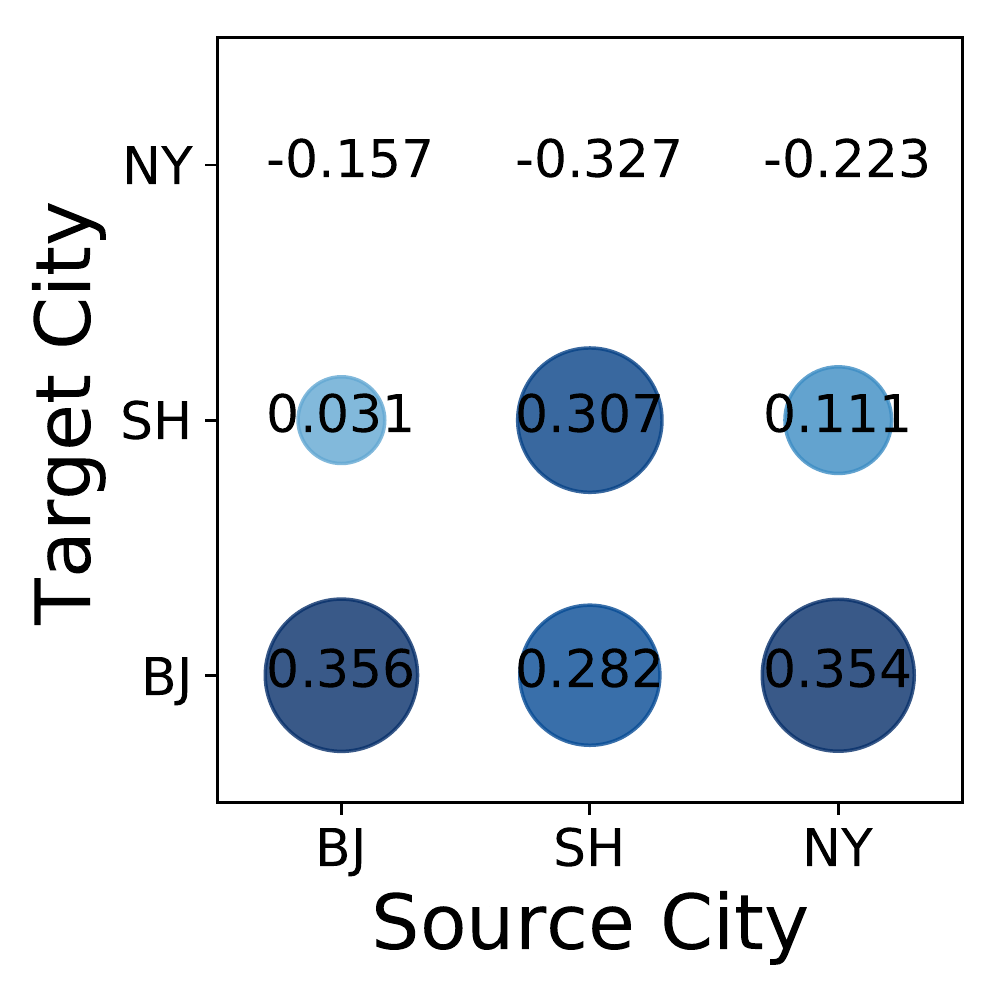}}
	\subfigure[Street View Imagery]{
		\includegraphics[width=0.2\textwidth]{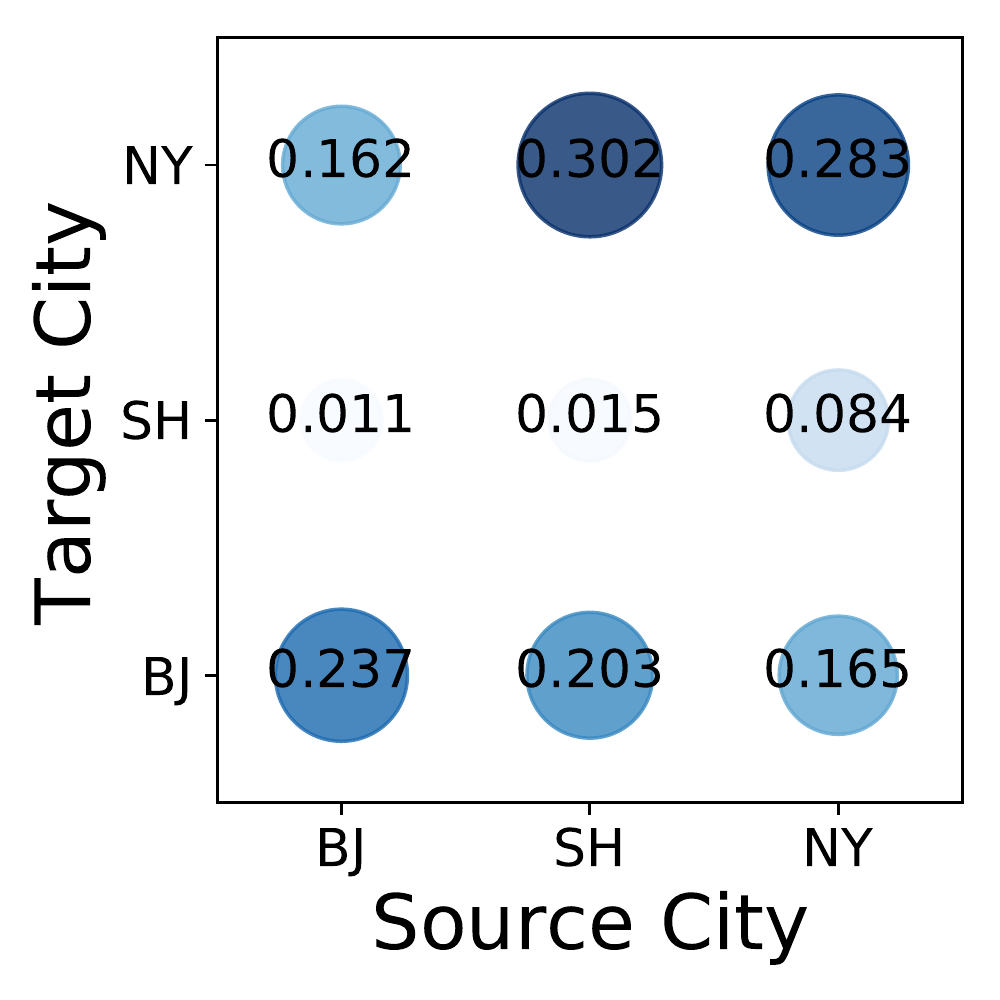}}
	\vspace{-10px}
	\caption{The $R^2$ for the transferability test of PG-SimCLR on satellite and street view imagery-based population prediction. The visual encoder is trained on one source city and then evaluated on other target cities.}\label{fig:transfer_simclr}
	\vspace{-10px}
\end{figure}

	\subsection{Data Sources \& Preprocessing}
	The data sources in our experiments and their links are provided as follows.
	\begin{itemize}[leftmargin=10px]
		\item \textbf{Satellite imagery data.} ArGIS, \url{https://geoenrich.arcgis.com/}.
		\item \textbf{Street view imagery data in Beijing \& Shanghai.} Baidu Map API, \url{http://api.map.baidu.com}.
		\item \textbf{Street view imagery data in New York.} Google Street API, \url{https://maps.googleapis.com/}.
		\item \textbf{Population data in Beijing \& Shanghai.} WorldPop. \url{https://www.worldpop.org/}.
		\item \textbf{Population \& education data in New York.} SafeGraph, \url{https://www.safegraph.com/}.
		\item \textbf{Crime data in New York.} NYC Open Data, \url{https://opendata.cityofnewyork.us/}.
	\end{itemize}
	 
	Since we focus on the more practical case of irregular region boundaries partitioned by road network, in the data preprocessing step, we merge multiple grid-based satellite images to match the region boundary. Thus, the satellite images for regions might be in irregular shape, as shown in Figure~\ref{fig:image_matching}(a).

	\section{Model Details \& Experiment Results} \label{sec:add_exp}
	\subsection{Training Algorithm}
	
	Algorithm~\ref{alg:1} summarizes the learning procedure of our proposed KnowCL model for urban imagery-based socioeconomic prediction. The overall framework is divided into two steps of knowledge-infused contrastive learning and socioeconomic prediction. In the first step, lines 4-5 build semantic encoder and visual encoder for different modalities of inputs, and lines 6-11 execute the cross-modality based contrastive learning in a minibatch way with model parameter updated. As for the second step in lines 12-14, only the regression module is trained with observed socioeconomic indicator data, which is then used for socioeconomic prediction.
	
	\algrenewcommand\algorithmicprocedure{\textbf{Step}}
	\newdimen{\algindent}
	\setlength\algindent{1.5em} %
	\algnewcommand\LeftComment[2]{%
		\hspace{#1\algindent}$\triangleright$ \eqparbox{}{#2} \hfill 
	}
	
	\begin{algorithm}
		
		\caption{Learning procedure of KnowCL model.}\label{alg:1}
		\begin{algorithmic}[1]
			
			\State \textbf{Input}: UrbanKG $\mathcal{G}=(\mathcal{E},\mathcal{R},\mathcal{F})$, urban imagery data $\mathcal{I}$, region set $\mathcal{A}$, socioeconomic indicator data $\mathcal{D}=\{(a,y_a)\vert a\in\mathcal{A}\}$.
			
			\State \textbf{Output}: The socioeconomic indicator $y_{a^\prime}$ for a region $a^\prime$ without observed label.
			
			\Procedure{1: Knowledge-infused Contrastive Learning}{}
			
			\State Initialize semantic encoder $f^{\text{KG}}(\cdot)$;
			\State Initialize visual encoder $f^{\text{Image}}(\cdot)$;
			
			\For{$i=1,2,\cdots,n_{\text{iter}}$}
			
			\State Sample a minibatch $\mathcal{A}_{\text{batch}}\in\mathcal{A}$ of size $m$;
			
			\State $\bm{e}_a=f^{\text{KG}}(\mathcal{G}, a),\bm{I}_a=f^{\text{Image}}(I_a),  \forall a \in \mathcal{A}_{\text{batch}}$; 
			
			\State $\tilde{\bm{e}}_a = g^{\text{KG}}(\bm{e}_a),  \tilde{\bm{I}}_a = g^{\text{Image}}(\bm{I}_a), \forall a \in \mathcal{A}_{\text{batch}}$;
			
			\State Compute the image-KG contrastive loss $\mathcal{L}_a$ using \eqref{eq:loss};
			
			\State Update encoder parameters w.r.t. the gradients, $\nabla\mathcal{L}_a$.
			
			\EndFor
			\EndProcedure
			
			\Procedure{2: Socioeconomic Prediction}{}
			\State Train regression module $\text{MLP}(\cdot)$ on $\mathcal{D}$;
			\State Predict socioeconomic indicator $y_{a^\prime}=\text{MLP}(f^{\text{Image}}(I_{a^\prime}))$.
			\EndProcedure
		\end{algorithmic}
	\end{algorithm}

	\subsection{Predicted v.s. True Indicators}
	Similar to the setting in Figure~\ref{fig:r2}, we present the comparison of predicted and true population results on Shanghai and New York in Figure~\ref{fig:r2_sh_ny}. It can be observed that most predicted samples are located along the line at 45\textdegree.

	\subsection{Transferability Study}
	To validate the transferability of our proposed KnowCL model, we also investigate the transferability of the best baseline PG-SimCLR in Figure~\ref{fig:transfer_simclr}. Compared with results of KnowCL in Figure~\ref{fig:transfer}, PG-SimCLR is less competitive in transfer setting.

	\subsection{Parameter Study}

	\begin{figure}[htbp]
		\centering
	\subfigure[Beijing]{
			\includegraphics[width=0.21\textwidth]{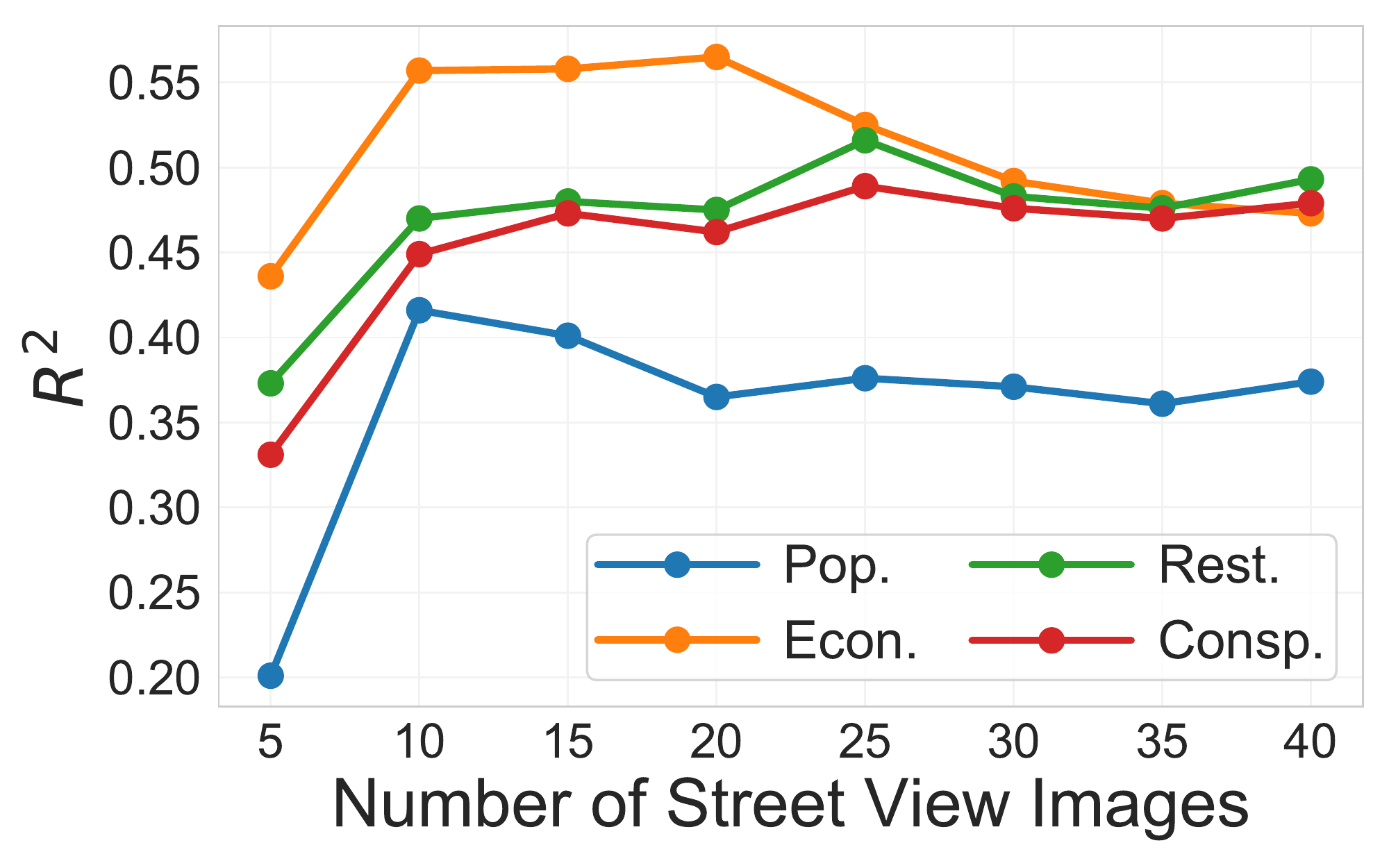}}
		\subfigure[New York]{
			\includegraphics[width=0.21\textwidth]{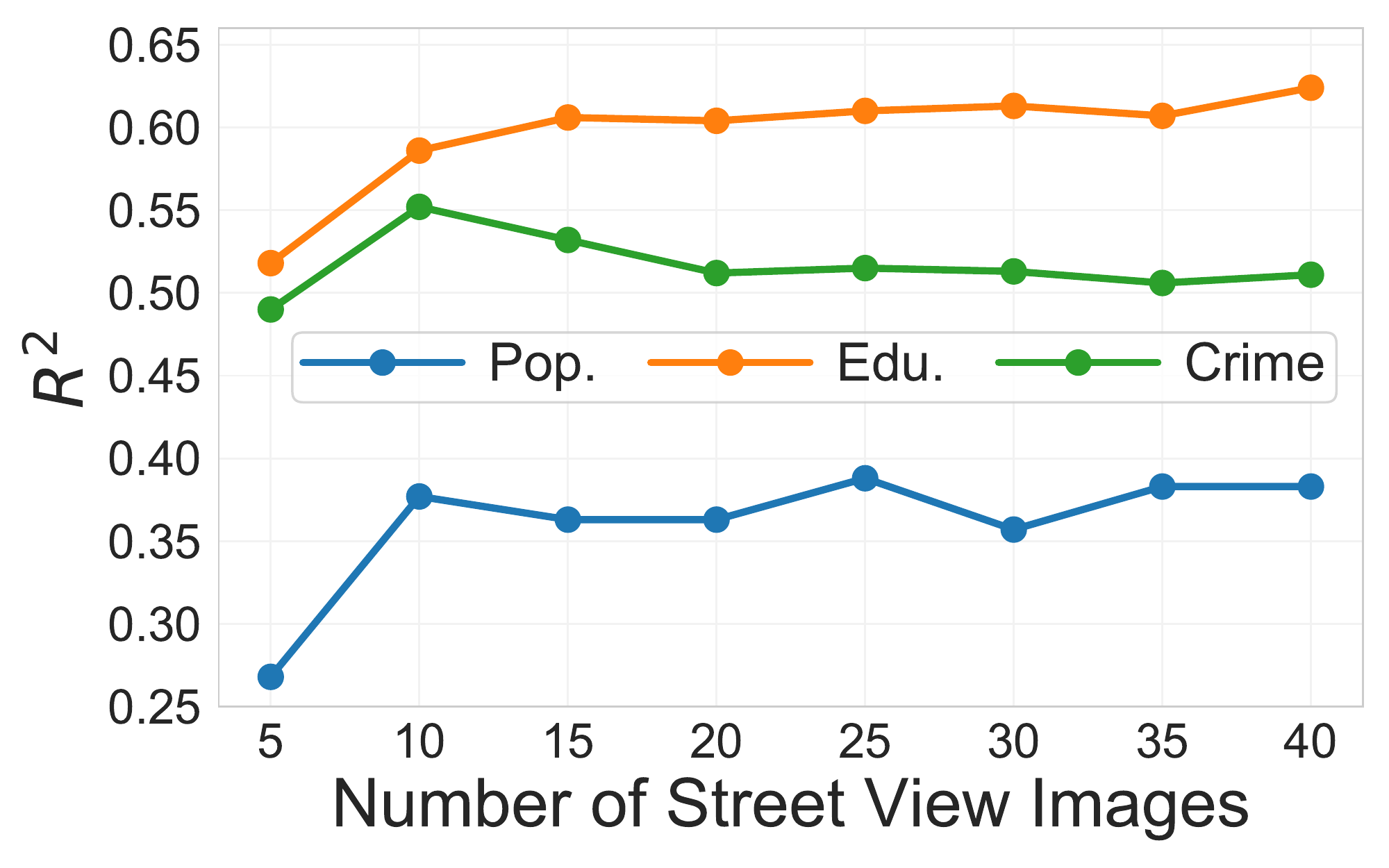}}
		\vspace{-10px}
		\caption{$R^2$ versus the number of street view images per region used for socioeconomic prediction on two datasets.}
		\label{fig:param1}
				\vspace{-15px}
	\end{figure}

	Figure~\ref{fig:param1} further investigates the influence of street view images on Beijing and New York datasets. Specifically, in the knowledge-infused contrastive learning step, we keep the number of street view images per region to 10, and tune the number of street view images used in socioeconomic prediction step. According to the results, with the increasing of used street view images, the prediction performance across most of socioeconomic indicators first increases and then converges. This phenomenon may partly owe to the image quality as well as the limited information captured in street view imagery. Moreover, such results also imply that introducing more stree view images for socioeconomic prediction may not bring performance improvement, which is heavily affected by the noise therein.

	\subsection{Component Analyses}

	To analyze the information captured in urban imagery representations, we employ principal component analysis (PCA) algorithm on learnt visual representations by KnowCL for dimension reduction, which are presented in Figure~\ref{fig:pca_vis}. We use the dot color to indicate the population at corresponding regions. Especially, the clustering phenomenon in respective of populations can be observed in Figure~\ref{fig:pca_vis}(a)-(d) on Beijing and Shanghai datasets, which validate the effectiveness of cross-modality based contrastive learning even without population supervision signals. As for the results in New York dataset, such phenomenon is not that obvious because the population in New York is uniformly distributed in block based regions, as we can see that most dots in Figure~\ref{fig:pca_vis}(e)-(f) are in similar colors. All the experiment results validate the effectiveness and robustness of our proposed knowledge-infused contrastive learning model for urban imagery-based socioeconomic prediction.

\end{document}